\documentclass[jlpea,article,accept,pdftex,moreauthors]{Definitions/mdpi}

\firstpage{1} 
\pubvolume{12}
\issuenum{2}
\articlenumber{28}
\pubyear{2022}
\copyrightyear{2022}
\externaleditor {\textls[-5]{{Academic Editor: Weidong Kuang}} 
}
\datereceived{10 March 2022} 
\dateaccepted{12 May 2022} 
\datepublished{18 May 2022} 
\hreflink{https://doi.org/10.3390/\\jlpea12020028} 

\usepackage{graphics}
\usepackage{caption}
\usepackage{listings}

\DeclareCaptionFormat{mylst}{#1#2#3}
\Title{Big--Little Adaptive Neural Networks on Low-Power Near-Subthreshold Processors}

\TitleCitation{Big--Little Adaptive Neural Networks on Low-Power Near-Subthreshold Processors}

\Author{Zichao Shen $^{1,}$*\orcidA{}, Neil Howard $^{2}$ and Jose Nunez-Yanez $^{1}$\orcidB{}}

\AuthorNames{Zichao Shen, Neil Howard and Jose Nunez-Yanez}

\AuthorCitation{Shen, Z.; Howard N.; Nunez-Yanez, J.}

\address{%
$^{1}$ \quad Electrical and Electronic Engineering, University of Bristol, Bristol, BS8 1UB, UK; \linebreak\{zs16916, j.l.nunez-yanez\}@bristol.ac.uk\\
$^{2}$ \quad Sensata Systems, Interface House, Swindon, SN4 8SY, UK; {njhoward.mobile@gmail.com}
}

\corres{Correspondence: zs16916@bristol.ac.uk}

\abstract{This paper investigates the energy savings that near-subthreshold processors can obtain in edge AI applications and proposes strategies to improve them while maintaining the accuracy of the application. The selected processors deploy adaptive voltage scaling techniques in which the frequency and voltage levels of the processor core are determined at the run-time. In these systems, embedded RAM and flash memory size is typically limited to less than 1 megabyte to save power. This limited memory imposes restrictions on the complexity of the neural networks model that can be mapped to these devices and the required trade-offs between accuracy and battery life. To address these issues, we propose and evaluate alternative `big--little' neural network strategies to improve battery life while maintaining prediction accuracy. The strategies are applied to a human activity recognition application selected as a demonstrator that shows that compared to the original network, the best configurations obtain an energy reduction measured at 80\% while maintaining the original level of inference accuracy.}


\begin{document}

\section{Introduction}
\label{Introduction}

Over the past few decades, the rapid development of the Internet of Things (IoT) and deep learning has increased the demand for deploying deep neural networks (DNNs) to low-power devices~\cite{zhou2019edge}. Due to high latency and privacy issues, cloud computing tasks are gradually being transferred to the edge in areas such as image recognition and natural language processing~\cite{chen2019deep}. The limitations in memory size and computing power mean that large neural networks with millions of parameters cannot be easily deployed on edge devices such as microcontroller units (MCUs), which in many cases have less than one megabyte of flash memory capacity~\cite{zhou2019edge, chen2019deep}. Memory is kept low to save costs and reduce power usage since power gating memory blocks that are not in use is not a feature available in these devices. 

Maximizing device usage time is an important goal and, focusing on this objective, we investigate an adaptive `big--little' neural network system which consists of a big network and multiple little networks to achieve energy-saving inference by limiting the number of big network executions without degrading accuracy. We call this organization `big--little' since it draws inspiration from the `big--little' technology popularized by ARM that combines complex and light processors in a single SoC. Our big network has better accuracy but with a longer inference time, while the little networks have a faster inference speed. Most of the time, the big network remains in sleeping mode and it is only activated when the little network determines that it cannot handle the work at the required level of confidence.

In this research, we focus on establishing and deploying the complete adaptive neural network system on the edge device. We investigate how to manage the primary and secondary networks to have a faster, more accurate, and more energy-efficient performance using a human activity recognition (HAR) application as a popular example of an edge application. The contribution of this research is summarized below:

\begin{itemize}
\item We evaluate state-of-the-art near-threshold processors with adaptive voltage scaling and compare them to a standard edge processor. 
\item We optimize a popular edge application targeting a human activity recognition (HAR) model based on \textit{TensorFlow} for MCU deployment using different vendor toolchains and compilers.
\item We propose novel `big--little' strategies suitable for adaptive neural network systems achieving fast inference and energy savings.
\item We made our work open source at \url{https://github.com/DarkSZChao/Big-Little_NN_Strategies} (accessed on 9 March 2022) to further promote work in this field.
\end{itemize}

This paper is organized as follows. In Section~\ref{Background and Related Work}, we present an overview of the state-of-the-art hardware for low-power edge AI, frameworks and relevant algorithmic techniques. Then, an initial evaluation in terms of performance and energy cost in near-threshold MCUs and standard MCUs was carried out in Section~\ref{Low-power Microcontroller Evaluation}. In Section~\ref{Adaptive Neural Network Methodology}, we propose and evaluate three different configurations of adaptive neural network systems with different features and performance characteristics. Section~\ref{Neural network Microcontroller Deployment} describes and demonstrates the implementation steps needed to target the selected low-power MCUs. The results obtained in terms of speed, accuracy and energy are presented in Section~\ref{Results and Discussion}. Finally, the conclusions and future work are discussed in Section~\ref{Conclusions}.

\section{Background and Related Work}
\label{Background and Related Work}

In this section, we present an overview of current state-of-the-art hardware with power profiles in the order of 1 watt or less for edge AI and then algorithmic techniques and frameworks optimized to target this hardware.

\subsection{Hardware for Low-Power Edge AI}
\label{Hardware for Low-power Edge AI}

The high demand for AI applications at the edge has resulted in a significant increase in hardware optimized for low-power levels. For example, Google has delivered a light version of the Tensor Processing Unit (TPU) called Edge TPU which is able to provide power-efficient inference at 2 trillion MAC operations per second per watt (2TMAC/s/W)~\cite{web_TPU}. This state-of-the-art device is able to execute mobile version models such as MobileNet V2 at almost 400~FPS. The Cloud TPU focuses on training complex models, while the Edge TPU is designed to perform inference in low-power systems. Targeting significantly lower power than the Edge TPU, Ambiq released the Apollo family of near-threshold processors based on the 32-bit ARM Cortex-M4F processor. These devices can reach much lower energy usage measured at only 6~\textmu{A/MHz} at 3.3~V under the working mode, and 1~\textmu{A/MHz} at 3.3~V under sleep mode. The Apollo3 device present in the SparkFun board has 1~MB of flash memory and 384~KB of low-leakage RAM~\cite{datasheet_Apollo3}. Similarly, Eta Compute has targeted energy-efficient endpoint AI solutions with the ECM3532 processor. This device is based on an ARM Cortex-M3 32-bit CPU and a separate CoolFlux DSP to speed up machine learning operations in an energy-efficient manner. The ECM3532 available in the AI vision board consumes less than 5~\textmu{A/MHz} in normal working mode and 1~\textmu{A/MHz} in sleep mode. According to Eta Compute, its implementation of self-timed continuous voltage and frequency scaling technology (CVFS) achieves a power profile of just 1~mW~\cite{datasheet_ECM3532, web_Eta}. A characteristic of these near-threshold devices is that voltage scaling is applied to the core but it is not applied to the device's SRAM/flash due to the limited margining possible in memory cells. 

Both Apollo3 and ECM3532 are based on the popular ARM architecture but, lately, the open-source instruction set architecture RISC-V has also received significant attention in this field. For example, GAP8 developed by GreenWaves Technologies features an 8-core compute cluster of RISC-V processors and an additional CNN accelerator~\cite{flamand2018gap}. The compute cluster is coupled with an additional ultra-low power MCU with 30~\textmu{W} state-retentive sleep power for control and communication functions. For CNN inference (90~MHz, 1.0~V), GAP8 delivers an energy efficiency of 600~GMAC/s/W and a worst-case power envelope of 75~mW~\cite{flamand2018gap}.

Other examples of companies exploring the near-threshold regime include Minima who has been involved in designs demonstrating achievable power savings~\cite{web_CEO}. Minima offers ultra-wide dynamic voltage and frequency scaling (DVFS) which is able to scale frequency and/or operating voltage based on the workload. This approach, combined with the dynamic margining approach from both Minima and ARM, is able to save energy by up to $15\times$ to $20\times$~\cite{web_Minima}. The interest for adaptive voltage scaling hardware has resulted in a \texteuro 100~m European project led by STMicroelectronics to develop the next generation of edge AI microcontrollers and software using low-power FD-SOI and phase change technology. This project aims to deliver the chipset and solutions for the automotive and industrial markets with a very high computing capacity of 10 TOPS per watt, which is significantly more powerful than existing microcontrollers~\cite{web_100m}. 

\subsection{Algorithmic Techniques for Low-Power Edge AI}
\label{Algorithmic Techniques for Low-power Edge AI}

Over the years, different algorithmic approaches have appeared to optimize inference on edge devices with a focus on techniques such as quantization, pruning, heterogeneous models and early termination. The deep quantization of network weights and activations is a well-known approach to optimize network models for edge deployments~\cite{novac2021quantization, jacob2018quantization}. Examples include~\cite{hubara2017quantized}, which uses extremely low precision (e.g., 1-bit or 2-bits) of weights and activations achieving 51\% top-1 accuracy and seven times the speedup in AlexNet~\cite{hubara2017quantized}. The authors of~\cite{courbariaux2016binarized} demonstrate a binarized neural network (BNN) where both weights and activations are binarized. During the forward pass, a BNN drastically reduces memory accesses and replaces most arithmetic operations with bit-wise operations. Ref.~\cite{courbariaux2016binarized} has proven that, by using their binary matrix multiplication kernel, the results achieve 32 times the compression ratio and improves performance by seven times with MNIST, CIFAR-10 and SVHN data sets. However, substantial accuracy loss (up to 28.7\%) has been observed by~\cite{rastegari2016xnor}. The research in~\cite{rastegari2016xnor} has addressed this drawback by deploying a full-precision norm layer before each Conv layer in XNOR-Net. XNOR-Net applies binary values to both inputs and convolutional layer weights and it is capable of reducing the computation workload by approximately 58 times, with 10\% accuracy loss in ImageNet~\cite{rastegari2016xnor}. Overall, these networks can free edge devices from the heavy workload caused by computations using integer numbers, but the loss of accuracy needs to be properly managed. This reduction in accuracy loss has been improved in CoopNet~\cite{mocerino2019coopnet}. Similar to the concept of multi-precision CNN in~\cite{amiri2018multi}, CoopNet~\cite{mocerino2019coopnet} applies two convolutional models: a binary net BNN with faster inference speed and an integer net INT8 with relatively high accuracy to balance the model's efficiency and accuracy. On low-power Cortex-M MCUs with limited RAM ($\leq$ 1~MB), Ref.~\cite{mocerino2019coopnet} achieved around three times the compression ratio and 60\% of the speed-up while maintaining an accuracy level higher than \linebreak the CIFAR-10, GSC and FER13 datasets. In contrast to CoopNet which applies the same network structures for primary and secondary networks, we apply a much simpler structure for secondary networks in which each of them is trained to identify one category in the HAR task. This optimization results in a configuration that can achieve around 80\% speed-up and energy-saving with a similar accuracy level across all the evaluated MCU platforms. Based on XNOR-Net, Ref.~\cite{romaszkan20203pxnet} constructed a pruned--permuted--packed network that combines binarization with sparsity to push model size reduction to very low limits. On the Nucleo platforms and Raspberry Pi, 3PXNet achieves a reduction in the model size by up to $38\times$ and an improvement in runtime and energy of $25\times$ compared to already compact conventional binarized implementations with a reduction in accuracy of less than 3\%. TF-Net is an alternative method that chooses ternary weights and four-bit inputs for DNN models. Ref.~\cite{yu2019tf} provides this configuration to achieve the optimal balance between model accuracy, computation performance, and energy efficiency on MCUs. They also address the issue that ternary weights and four-bit inputs cannot be directly accessed due to memory being byte-addressable by unpacking these values from the bitstreams before computation. On the STM32 Nucleo-F411RE MCU with an ARM Cortex-M4, Ref.~\cite{yu2019tf} achieved improvements in computation performance and energy efficiency of $1.83\times$ and $2.28\times$, respectively. Thus, 3PXNet/TF-Net can be considered orthogonal to our `big--little' research since they could be used as alternatives to the 8-bit integer models considered in this research. A related architecture to our approach called BranchyNet with early exiting was proposed in~\cite{teerapittayanon2016branchynet}. This architecture has multiple exits to reduce layer-by-layer weight computation and I/O costs, leading to fast inference speed and energy saving. However, due to the existence of multiple branches, it suffers from a huge number of parameters, which would significantly increase the memory requirements in edge devices. 

The configuration of primary and secondary neural networks has been proposed for accelerating the inference process on edge devices in recent years. Ref.~\cite{amiri2018multi, park2015big} constructed `big' and `little' networks with the same input and output data structure. The `big' network is triggered by their score metric generated from the `little' network. A similar configuration has also been proposed by~\cite{nunez2021energy}, but their `big' and `little' networks are trained independently. `Big' and `little' networks do not share the same input and output data structure. Ref.~\cite{nunez2021energy} proposed a heterogeneous setup deploying a `big' network on state-of-the-art edge neural accelerators such as NCS2, with a `little' network on near-threshold processors such as ECM3531 and Apollo3. Ref.~\cite{nunez2021energy} has successfully achieved 93\% accuracy and low energy consumption of around 4~J on human activity classification tasks by switching this heterogeneous system between `big' and `little' networks. Ref.~\cite{nunez2021energy} considers heterogeneous hardware, whereas our approach uses the `big--little' concept but focuses on deploying all the models on a single MCU device. In contrast to how~\cite{nunez2021energy} deployed `big' and `little' models on the NCS2 hardware accelerator and near-threshold processors separately, we deploy both neural network models on near-threshold MCU for activity classification tasks. A switching algorithm is set up to switch between `big' and `little' network models to achieve much lower energy costs but maintain a similar accuracy level. A related work~\cite{ordonez2016deep} has performed activity recognition tasks with excellent accuracy and performance by using both convolutional and long short-term memory (LSTM) layers. Due to the flash memory size of MCU, we decided not to use the LSTM layers which have millions of parameters as shown in~\cite{ordonez2016deep}. The proposed adaptive system is suitable for real-world tasks such as human activity classification in which activities do not change at very high speeds. A person keeps performing one action for a period of time, typically in the order of tens of seconds~\cite{turaga2008machine}, which means that to maintain the system at full capacity (using the primary `big' network to perform the inference) is unnecessary. Due to the additional inference time and computation consumed by the primary network, the fewer the number of times the primary network gets invoked, the faster the inference process will be and the lower the energy requirements~\cite{nunez2021energy, mocerino2019coopnet, amiri2018multi, park2015big}.

\subsection{Frameworks for Low-Power Edge AI}
\label{Frameworks for Low-power Edge AI}

Over the last few years, a number of frameworks have appeared to ease the deployment of neural network models on edge devices with limited resources. In~\cite{wang2020fann}, a framework is provided called FANN-on-MCU specifically for the fast deployment of multi-layer perceptrons (MLPs) on low-power MCUs. This framework supports not only the very popular ARM Cortex-M series MCUs, but also the RISC-V parallel ultra-low power (PULP) processors. The results in~\cite{wang2020fann} show that the PULP-based `Mr.Wolf' SoC can reach up to $7.1\times$ the speedup with respect to a single core implementation and $13.5\times$ the speedup over the ARM Cortex-M4. Moreover, by using FANN-on-MCU, a relatively big neural network with 103,800 MAC operations can be executed within 17.6~ms with an energy consumption of 183~\textmu{J} on a Nordic nRF52832 MCU with one ARM Cortex-M4. The same neural network applied on `Mr.Wolf' with eight RISC-V-based RI5CY cores takes less than 1ms to consume around 50~\textmu{J}~\cite{wang2020fann}. Similar to FANN-on-MCU, Ref.~\cite{web_NNoM} delivers a fast deployment on the MCU framework called the neural network on microcontroller (\textit{NNoM}) which supports more complex model topologies such as ResNet and DenseNet from Keras. A user-friendly API and high-performance backend selections have been built for embedded developers to deploy Keras models on low-power MCU devices. There are also deployment frameworks developed by commercial companies targeting low-power edge devices. For example, Google focuses on low-power edge AI with the popular \textit{TensorFlow Lite} 
framework~\cite{web_TFLite}. Coupled with the model training framework \textit{TensorFlow}, Google can provide a single solution from neural network model training to model deployment on edge devices. \textit{STM32Cube.AI} from STMicroelectronics~\cite{web_STM32} is also an AI deployment framework but it is only designed around the STM family devices such as STM32 Nucleo-L4R5ZI and STM32 Nucleo-F411RE. Eta Compute has created the \textit{TENSAIFlow} deployment framework to provide performance and efficiency optimizations for Eta-series MCU products such as ECM3531 and ECM3532~\cite{web_TENSAI}. In our methodology, the lack of support for certain devices in some frameworks means that we have combined tools from different vendors. We have applied frameworks \linebreak from~\cite{web_NNoM, web_TFLite, web_TENSAI} for model deployments on MCUs such as ECM3532 and STM32L4 (see Section~\ref{Neural network Microcontroller Deployment} for details).

\section{Low-Power Microcontroller Evaluation}
\label{Low-power Microcontroller Evaluation}

Four commercially available microcontroller devices designed for energy-efficient applications from STMicroelectronics, Ambiq and Eta Compute are considered in this comparison. Table~\ref{MCU} shows the technical details of these four MCUs. Three of them (STM32L4R5ZI, Apollo2 Blue and SparkFun Edge (Apollo3 Blue)) are based on the Cortex-M4 microarchitecture with floating-point units (FPU)~\cite{datasheet_Apollo3, datasheet_STM32, datasheet_Apollo2}, while the ECM3532 is based on the Cortex-M3 microarchitecture with a `CoolFlux' 16-bit DSP~\cite{datasheet_ECM3532}. The 32-bit ARM Cortex-M3 and M4 are comparable microarchitectures both having a three-stage pipeline and implementing the Thumb-2 instruction set with some differences in the number of instructions available. For example, additional 16/32-bit MAC instructions and single-precision FPU are only available on the Cortex M4.

The STM32 Nucleo-144 development board with the STM32L4R5ZI MCU is used as a comparison point; the main difference between this STM device and the other three is the power optimization method. The core supply voltage of 1~V for the STM device is significantly higher than the core voltage for the near-threshold devices of Ambiq and Eta Compute at only around 0.5~V. Theoretically, the sub-threshold core supply voltage can be as low as 0.3~V which should be more power-efficient. However, at 0.3~V, the transistor switching time will be longer, which leads to a higher leakage current. The leakage can exceed 50\% of the total power consumption for a threshold voltage level of around 0.2~V~\cite{yeo2004low}. Therefore, in practice, choosing near-threshold voltage points instead of sub-threshold voltage points has been shown to be a more energy-efficient solution~\cite{yeo2004low}. In order to optimize the energy usage based on the task requirements, STM32L4 uses standard dynamic voltage and frequency scaling (DVFS) with predefined pair sets of voltage and frequency, while the devices from Ambiq and Eta Compute apply adaptive voltage scaling (AVS) which is able to determine the voltage at a given frequency to handle the tasks at run-time using a feedback loop~\cite{nunez2018energy}.

\begin{figure}[H]
\centering
\includegraphics[width=22.5cm]{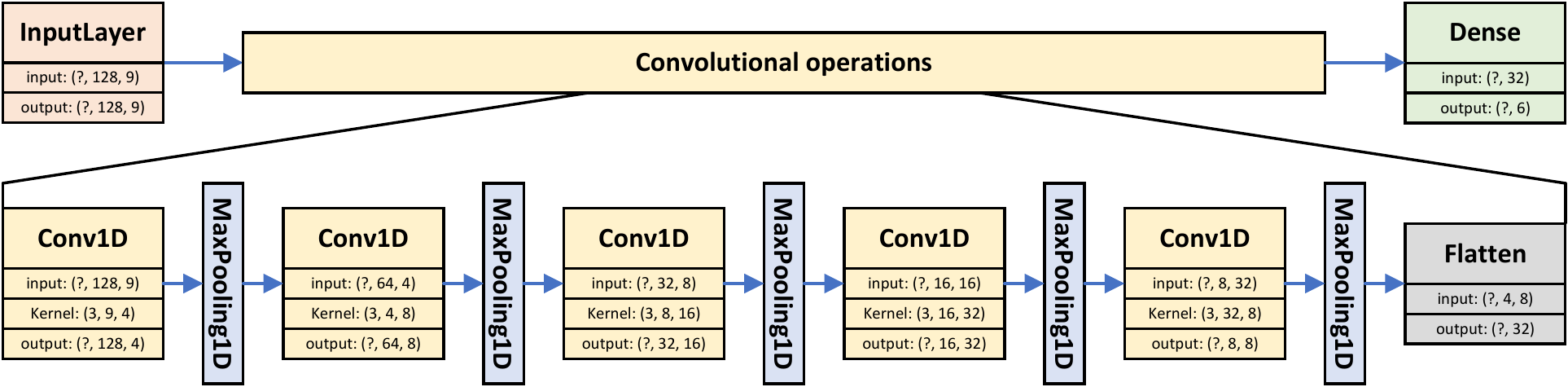}
\caption{Convolutional neural network for the initial performance and energy evaluation of the MCU.\label{NNStruc_B_1IN}}
\end{figure}

Comparing the datasheets, the STM32L4 has the highest clock frequency which results in an advantage in processing speed. Ambiq and Eta Compute's near-threshold devices only require about half of the core supply voltage of STM32L4. All considered processors are equipped with limited flash sizes from 0.5~MB to 1~MB and a size of around 300~KB SRAM. That means that the neural network model deployed must be small enough to fit within the limited memory size. Therefore, we use the \textit{TensorFlow} framework and \textit{TensorFlow Lite} converter to create a simple pre-trained CNN model designed for human activity recognition (HAR) from UCI~\cite{web_UCI} (as shown in Figure~\ref{NNStruc_B_1IN}) to perform the initial energy evaluation of the four MCU devices.

\startlandscape
\begin{table}[H]
\caption{Low-power MCU comparison.\label{MCU}}
	\newcolumntype{C}{>{\centering\arraybackslash}X}
\begin{tabularx}{\textwidth}{CCCm{3cm}<{\centering}m{2.5cm}<{\centering}CCCm{3.8cm}<{\centering}}
			\toprule
			\textbf{Devices} & \textbf{Manufacturer} & \textbf{Architecture} & \textbf{Embedded Memory (Flash/SRAM/Cache)} & \textbf{Clock Frequency} & \textbf{Core Voltage} & \textbf{Sleeping Mode Current} & \textbf{Work Mode Current} & \textbf{Power Scaling Capabilities}\\
			\midrule
			STM32L4R5ZI &  STMicroelectronics &  32-bit Cortex-M4 CPU with FPU &  1~MB/320~KB &  up to 120~MHz &  1.05~V &  $\textless$5~\textmu{A} with RTC &  110~\textmu{A/MHz} &  Dynamic voltage scaling with two main voltage ranges\\
	
			Apollo2 Blue &  Ambiq &  32-bit Cortex-M4 CPU with FPU &  1~MB/256~KB/16~KB &  up to 48~MHz &  0.5~V &  $\textless$3~\textmu{A} with RTC &  10~\textmu{A/MHz} &  SPOT (Subthreshold Power Optimized Technology)\\
			
			SparkFun Edge (Apollo3 Blue) &  Ambiq &  32-bit Cortex-M4 CPU with FPU &  1~MB/384~KB/16~KB &  up to 96~MHz (with TurboSPOT Mode) &  0.5~V &  1~\textmu{A} with RTC &  6~\textmu{A/MHz} &  SPOT (Subthreshold Power Optimized Technology) with TurboSPOT\\
			
			ECM3532 &  Eta Compute &  32-bit Cortex-M3 CPU with 16-bit `CoolFlux' DSP	&  512~KB/256~KB/0~KB &  up to 100~MHz &  0.55~V &  1~\textmu{A} with RTC &  5~\textmu{A/MHz} &  CVFS (Continuous Voltage Frequency Scaling) with multiple frequency and voltage points\\
			\bottomrule
		\end{tabularx}
\end{table}

\vspace{-6pt}

\finishlandscape

\begin{figure}[H]
\includegraphics[width=\textwidth]{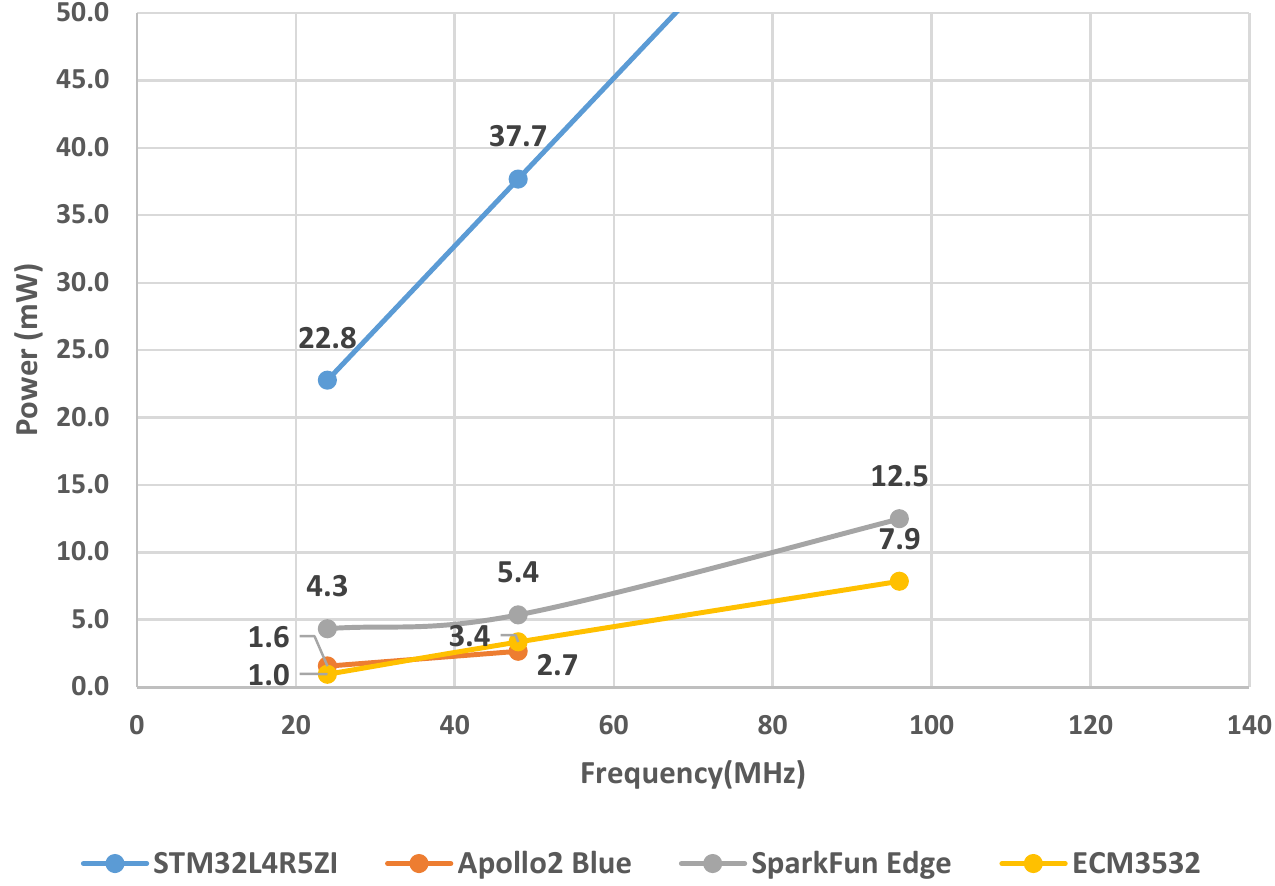}
\sethlcolor{red}
\caption{MCU initial evaluation in terms of power consumption.\label{Initial_Power}}
\end{figure}

The energy board X-NUCLEO-LPM01A from STMicroelectronics is used to evaluate the performance and energy consumption measuring the current used by the target board under a given supply voltage of 3.3~V (lower core voltages are regulated internally in the device). The power consumption of the four tested boards is shown in Figure~\ref{Initial_Power}. STM32L4 operates at a much higher power level which is around six times that of the near-threshold processors. The near-threshold processors Apollo2, Apollo3 and ECM offer significantly lower power, consuming less than 5~mW at the normal frequency of 48MHz and around 10~mW in the burst mode of 96~MHz. The reason why SparkFun Edge (Apollo3) consumes more power than Apollo2 is that the Apollo3 core is highly integrated into the SparkFun Edge board with peripheral sensors and ports which cannot be disabled during the power evaluation. Therefore, the peripheral devices on SparkFun Edge (Apollo3) are responsible for a component of the power consumption, which leads to a higher power than Apollo2 at each frequency level. Apollo2 and ECM3532 share a similar level of power consumption at 24 and 48~MHz. Apollo2 does not support running at a frequency higher than 48~MHz; therefore, there is no value for Apollo2 at the 96~MHz frequency point. 

\begin{figure}[H]
\includegraphics[width=\textwidth]{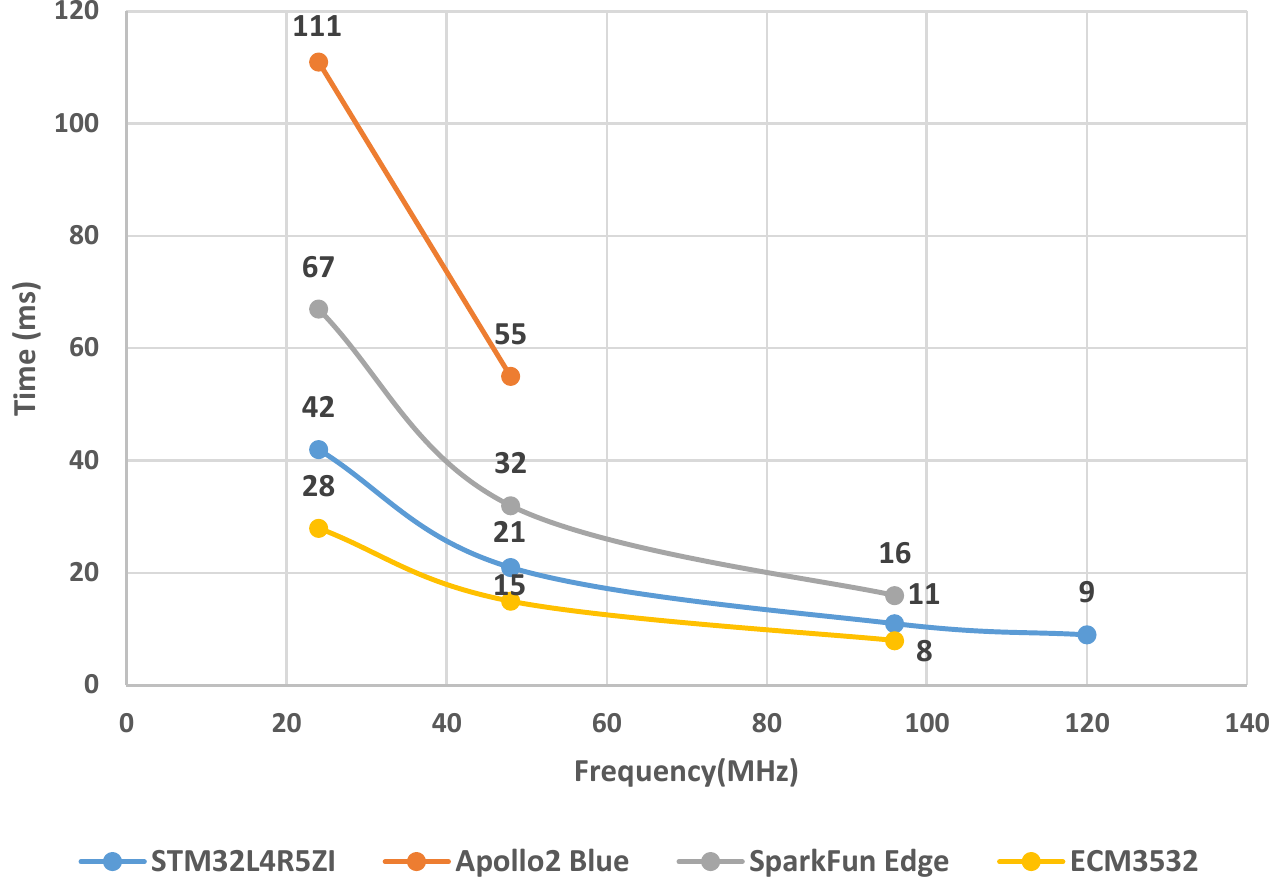}
\caption{MCU initial evaluation in terms of time cost.\label{Initial_Time}}
\end{figure}

Figure~\ref{Initial_Time} shows the execution time of the four tested processors for one inference of the pre-trained CNN model in Figure~\ref{NNStruc_B_1IN}. Apollo2 is the slowest one and finishes inference using the longest amount of time at above 100~ms at 24~MHz frequency and around 50~ms at 48~MHz. The SparkFun Edge board (Apollo3) reduces the execution time by approximately 40\% compared to Apollo2. It can even drop below 20~ms when operating in burst mode (96~MHz). STM32L4 is the second fastest among all devices due to its higher core supply voltage in Table~\ref{MCU} which enables faster transistor switching and processing speed. ECM3532 has the lowest execution times which are 28~ms at 24~MHz, 15~ms at 48~MHz and 8~ms at 96~MHz. The \textit{TENSAIFlow} compiler is responsible for significant optimization in the ECM3532 device.

\begin{figure}[H]
\includegraphics[width=\textwidth]{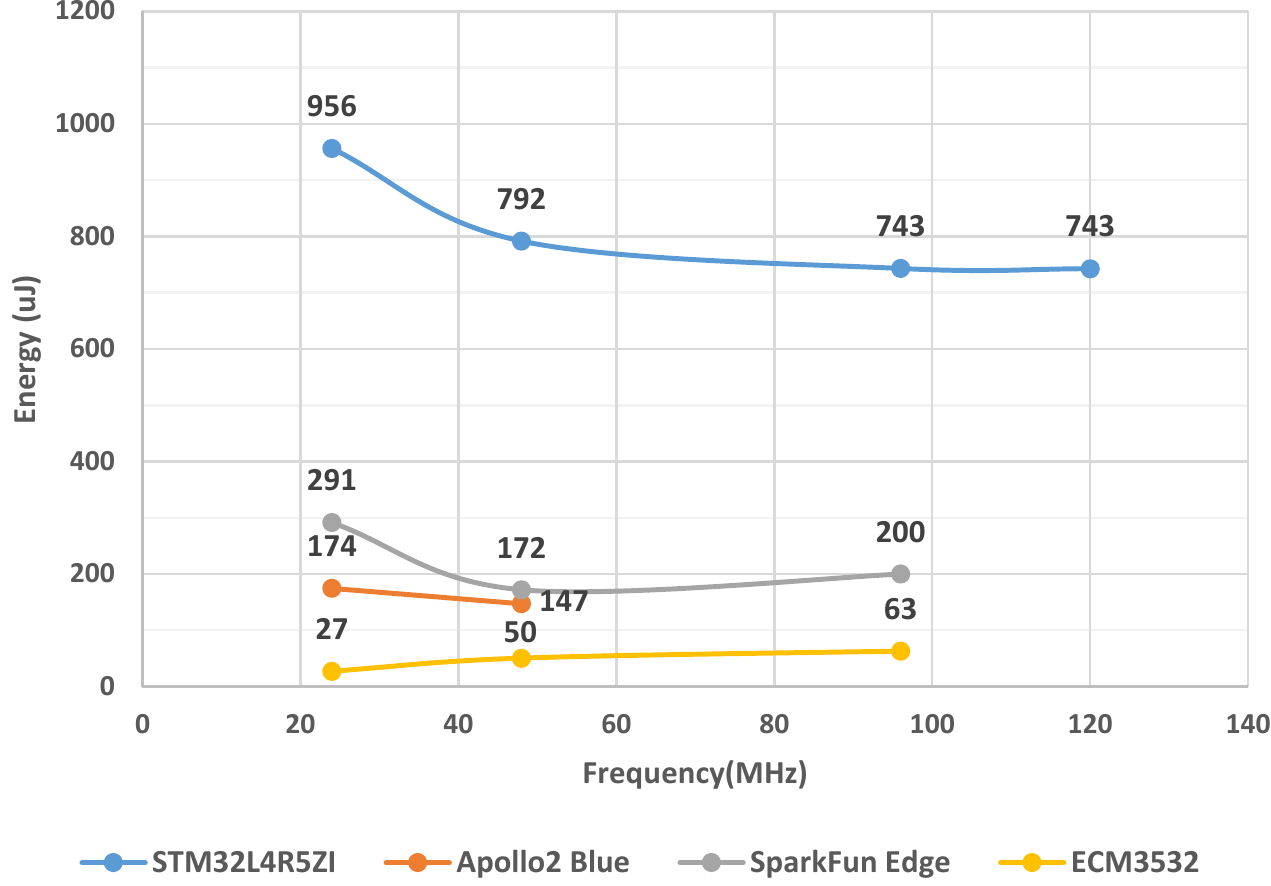}
\caption{MCU initial evaluation in terms of energy consumption.\label{Initial_Energy}}
\end{figure}

Figure~\ref{Initial_Energy} indicates the energy consumption values observed using the X-NUCLEO-LPM01A energy measurement board. Since the power consumption of the standard MCU STM32L4 in Figure~\ref{Initial_Power} is six times higher compared to the near-threshold MCUs and there is no obvious advantage in processing speed at the same frequency, STM32L4 is the worst device in terms of energy consumption for all operating frequencies from 24 to 96~MHz. SparkFun Edge (Apollo3) is slightly higher than Apollo2 at 24 and 48MHz due to the energy consumed by the peripheral equipment on board. ECM3532 achieves the minimum energy consumption at normal frequency points (24 and 48~MHz) in the energy test because it has better results in both power and time evaluations. However, when operating in the 96~MHz burst mode, ECM3532 requires more power to obtain a higher processing speed, resulting in a slight increase in energy consumption, and the same situation can be seen for the SparkFun Edge board.

Overall, compared to the STM32L4 reference point all three near-threshold MCUs have a significant advantage in power and energy consumption which is around 80\% to 85\% lower. Although the near-threshold MCUs are comparable with the standard MCU STM32L4 in terms of inference time, their lower core voltage supplies (Table~\ref{MCU}) result in lower power (Figure~\ref{Initial_Power}) at the same frequency level. Therefore, in our model inference evaluation, the near-threshold MCU devices can achieve better results in energy consumption compared to STM32L4 at 24, 48 and 96~MHz. Thanks to the additional model optimization obtained with the \textit{TENSAIFlow} compiler provided by Eta Compute, ECM3532 offers a good balance between performance and energy efficiency to reach a lower execution time, enabling the lowest energy consumption for model inference from 24 to 96~MHz. In contrast, Apollo2, with a relatively slow processing speed, needs more time for model inference, which leads to higher values in energy consumption at 24 and 48~MHz. Due to the energy consumed by the inaccessible peripheral equipment on SparkFun Edge (Apollo3), this device consumes higher energy than Apollo2 (Figure~\ref{Initial_Energy}). 

\unskip

\unskip

\section{Adaptive Neural Network Methodology}
\label{Adaptive Neural Network Methodology}

To create the adaptive neural network system, we employ Python version 3.6.8 and \textit{TensorFlow} 1.15 with its dependencies installed on a desktop PC with Intel(R) Core (TM) i7-10850H CPU \@ 2.70~GHz, NVIDIA GeForce MX250 GPU, and 16~GB RAM. There are several framework alternatives to train the neural networks, such as PyTorch and Caffe. Due to the reasons of MCU compatibility and stability, our approach uses \textit{TensorFlow} 1.15 to train the primary and secondary network models. After that, we use \textit{TensorFlow Lite} and \textit{NNoM} Converter to convert the models using single-precision floating-points (FP32) to the unsigned integer 8-bit (UINT8) format which can be deployed on the MCUs.

We consider human activity recognition using the UCI data set~\cite{web_UCI} as our raw data set. This application is a demonstrator which assumes that the activity will remain constant for a short period of time before being replaced by the next activity. To save energy via a reduction in execution time, we propose the adaptive neural network system which is able to disable the primary model and activate a secondary model when the activity remains unchanged. Therefore, we aim at achieving both latency and energy reductions without affecting prediction accuracy.

\begin{figure}[H]
\includegraphics[width=\textwidth]{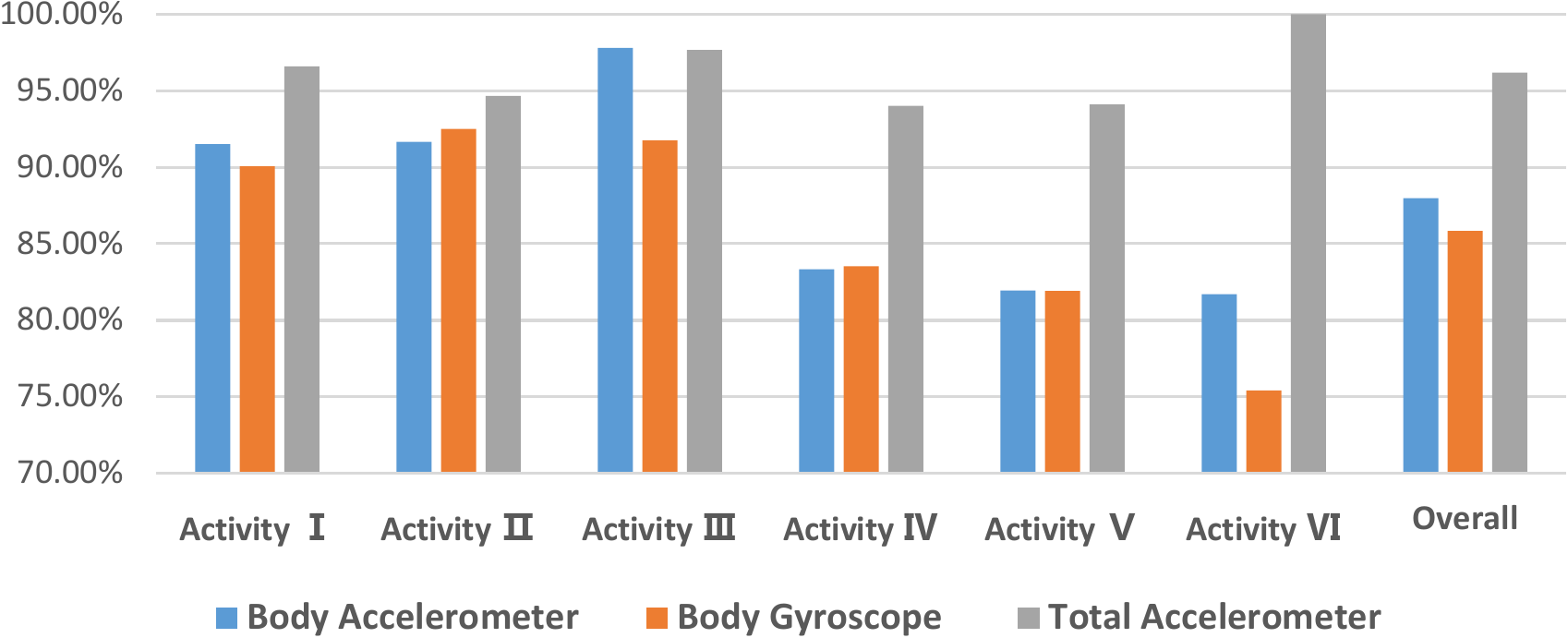}
\caption{The test accuracy of three sensors on the secondary network. The total accelerometer sensor which achieves the highest overall accuracy is chosen for the secondary network.\label{s123_Accuracy}}
\end{figure}

The UCI-HAR data set uses a body accelerometer, body gyroscope, and total accelerometer with three axes to provide body information for six actions (SITTING, STANDING, LAYING, WALKING, WALKING\_UPSTAIRS, and WALKING\_DOWNSTAIRS) performed by a group of 30 volunteers. All the data have been sampled in fixed-width sliding windows of 128 sampling points and they have been randomly partitioned into two sets, with 70\% of data samples used for training and 30\% used for testing. Therefore, we have a training data shape of (7352, 128, 3, 3), and a testing data shape of (2947, 128, 3, 3). We have evaluated the accuracy as shown in Figure~\ref{s123_Accuracy} by applying the test data from these three sensors to the secondary network. The total accelerometer sensor shows the best overall accuracy. Thus, this sensor is selected for the secondary network inference. The training and testing data sets from UCI-HAR use floating-point values with a wide range so that before training the model, all the data have been rescaled to quantized integer values with a range of [\textminus128, 127]. In the following sections, these actions are referred to as follows:

\begin{itemize}
\item Activity \uppercase\expandafter{\romannumeral1} = WALKING
\item Activity \uppercase\expandafter{\romannumeral2} = WALKING\_UPSTAIRS
\item Activity \uppercase\expandafter{\romannumeral3} = WALKING\_DOWNSTAIRS
\item Activity \uppercase\expandafter{\romannumeral4} = SITTING
\item Activity \uppercase\expandafter{\romannumeral5} = STANDING
\item Activity \uppercase\expandafter{\romannumeral6} = LAYING
\end{itemize}

In principle, if the system uses the `big'-only configuration, then the accuracy will be higher. The challenge is to reduce the `big' activation count without reducing accuracy. To achieve this, we propose the three alternative configurations shown below:

\begin{itemize}
\item `Big'-only (original method)
\item `Big' + six `little'
\item `Big' + `dual'
\item `Big' + distance.
\end{itemize}

\subsection{`Big' + Six `Little' Configuration}
\label{`Big' + Six `Little' Configuration}

\begin{figure}[H]
\begin{adjustwidth}{-\extralength}{0cm}
\centering
\includegraphics[width=17.5cm]{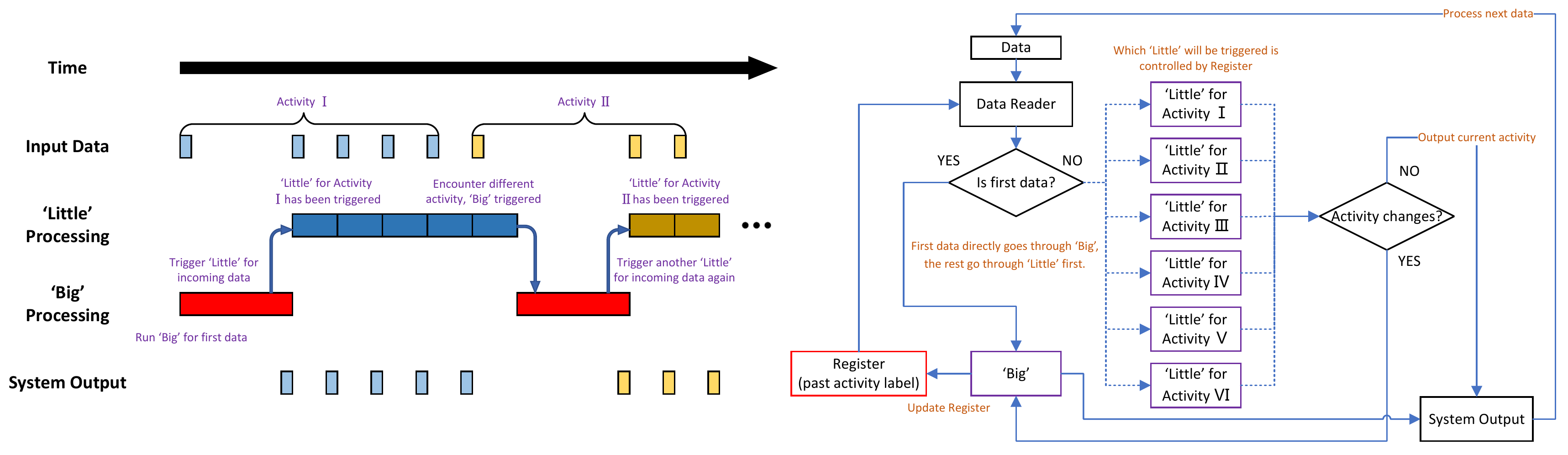}
\end{adjustwidth}
\caption{The processing steps (\textbf{left}) and the flow chart (\textbf{right}) for the `big' + six `little' configuration of the adaptive neural network system. In the left figure, dark blue and brown represent two `little' network models corresponding to the input activities. In the right figure, the dotted line means only one `little' network model of six is invoked at a time.\label{Config_1B6S}}
\end{figure}

The main idea is to deploy an adaptive system that consists of one `big' (output of six classes) and six `little' (output of two classes for each) network models in the MCU devices. As shown in Figure~\ref{Config_1B6S}, once the `big' network provides a category and stores it in the register, the corresponding `little' network will be triggered to infer the next incoming activity. The `little' network is responsible for detecting whether this current activity belongs to the previous category provided by the `big' network. For example, after the first activity is detected as SITTING by triggering the `big' network, the second incoming activity data will be processed by the `little' network, which is able to distinguish whether the second one is still SITTING or not. Once the next incoming activity is not SITTING anymore, the `big' network will be switched back on again. Otherwise, the `little' network remains active to save power and inference time. Therefore, we have six `little' networks which are equal to the number of categories classified in the `big' network.

\begin{figure}[H]
\begin{adjustwidth}{-\extralength}{0cm}
\centering
\includegraphics[width=17.5cm]{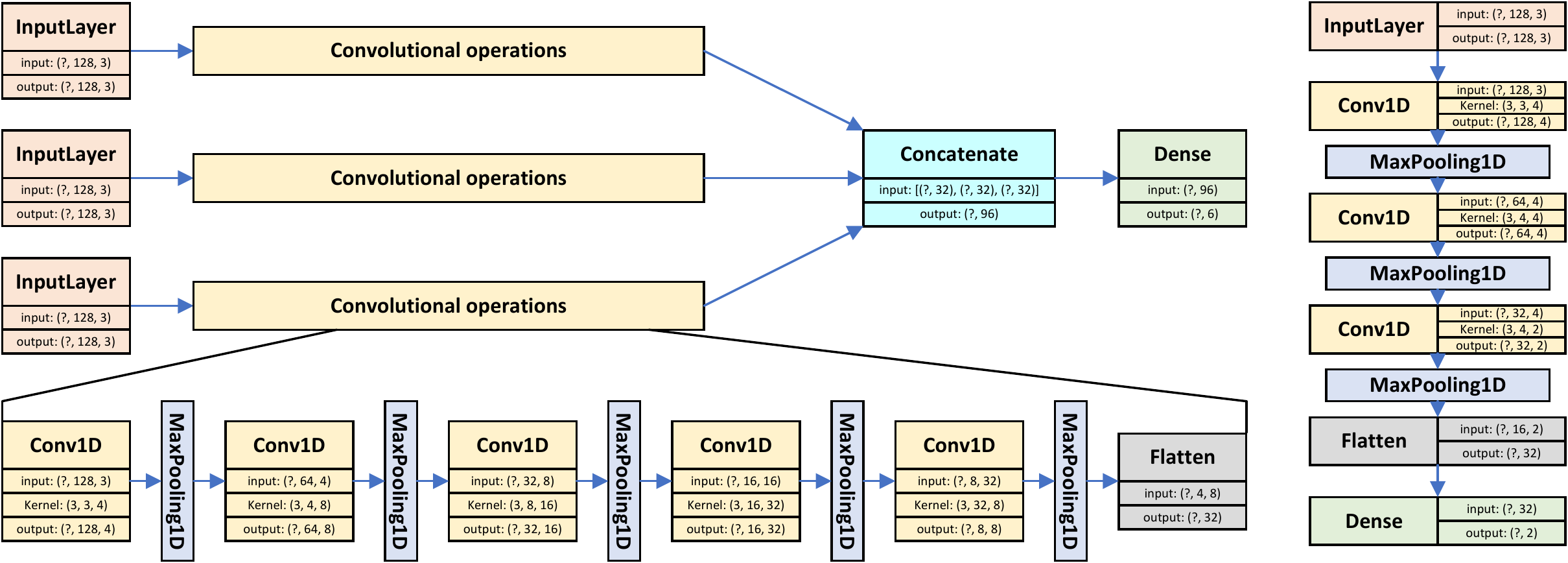}
\end{adjustwidth}
\caption{`Big' (left) and `little' (right) model structures in Keras.\label{NNStruc_B_3IN_S}}
\end{figure}

Figure~\ref{NNStruc_B_3IN_S} shows the model topologies for both `big' and `little' networks, while Table~\ref{Param_BS} shows the parameter details of the network layers. Since we are using all three sensors' data from the UCI-HAR data set to classify six activities, the `big' network has three inputs, resulting in around 9000 parameters in total. Convolutional 1D layers and max-pooling layers from Keras are stacked together to form the three `big' branches in Figure~\ref{NNStruc_B_3IN_S}. Then, the outputs from these branches are converged by a concatenate layer followed by a dense layer that has six neurons for six categories. The data shape of each sensor is (7352, 128, 3) which means we have 7352 data samples with a length of 128 for each axis. The data set is labelled from 0 to 5 to represent each activity for the training and testing processes in the `big' network.

\vspace{-6pt}

Each `little' network only classifies two categories by using several convolutional 1D layers and max-pooling layers with 184 parameters in total. Therefore, based on the results in Figure~\ref{s123_Accuracy}, only the total accelerometer sensor which achieves the best overall accuracy is selected as the input for the `little' network. The output of the `little' network is a dense layer with two neurons for two categories as seen in Table~\ref{Param_BS}. Due to the limited size of the UCI-HAR data set~\cite{web_UCI}, we have less than 2000 data elements for each activity category. Therefore, we use all of them and convert the data labels from six categories to two for training the `little' model. Particularly, for each `little' model, the labels of corresponding activity are set to number~1, while the others are set to number 0. Finally, we can generate the models in the Keras format.

\begin{table}[H]
\caption{`Big' (\textbf{left}) and `little' (\textbf{right}) model parameter details. The pooling layers are hidden. For more info, see Figure~\ref{NNStruc_B_3IN_S}. \label{Param_BS}}
    \begin{tabularx}{\textwidth}{m{2cm}<{\centering}Cm{1cm}<{\centering}m{2cm}<{\centering}Cm{1cm}<{\centering}}
        \toprule
        \multicolumn{3}{c}{\textbf{Model: `Big'}} & \multicolumn{3}{c}{\textbf{Model: `Little'}}\\
        \midrule
        \textbf{Layer (Type)} & \textbf{Output Shape} & \textbf{Param\#} & \textbf{Layer (Type)} & \textbf{Output Shape} & \textbf{Param\#}\\
        \midrule
        model\_input1 & [(None, 128, 3)] & 0 & model\_input & [(None, 128, 3)] & 0\\ 
        model\_input2 & [(None, 128, 3)] & 0 & conv1d & (None, 128, 4) & 40\\ 
        model\_input3 & [(None, 128, 3)] & 0 & conv1d\_1 & (None, 64, 4) & 52\\ 
        conv1d & (None, 128, 4) & 40 & conv1d\_2 & (None, 32, 2) & 26\\ 
        conv1d\_5 & (None, 128, 4) & 40 & model\_output & (None, 2) & 66\\ 
        conv1d\_10 & (None, 128, 4) & 40 & ~ & ~ &\\ 
        conv1d\_1 & (None, 64, 8) & 104 & ~ & ~ &\\ 
        conv1d\_6 & (None, 64, 8) & 104 & ~ & ~ &\\ 
        conv1d\_11 & (None, 64, 8) & 104 & ~ & ~ &\\ 
        conv1d\_2 & (None, 32, 16) & 400 & ~ & ~ &\\ 
        conv1d\_7 & (None, 32, 16) & 400 & ~ & ~ &\\ 
        conv1d\_12 & (None, 32, 16) & 400 & ~ & ~ &\\ 
        conv1d\_3 & (None, 16, 32) & 1568 & ~ & ~ &\\ 
        conv1d\_8 & (None, 16, 32) & 1568 & ~ & ~ &\\ 
        conv1d\_13 & (None, 16, 32) & 1568 & ~ & ~ &\\ 
        conv1d\_4 & (None, 8, 8) & 776 & ~ & ~ &\\ 
        conv1d\_9 & (None, 8, 8) & 776 & ~ & ~ &\\ 
        conv1d\_14 & (None, 8, 8) & 776 & ~ & ~ &\\ 
        concatenate & (None, 96) & 0 & ~ & ~ &\\ 
        model\_output & (None, 6) & 582 & ~ & ~ &\\
        \midrule
        \multicolumn{3}{c}{Total params: 9246} & \multicolumn{3}{c}{Total params: 184}\\
        \bottomrule
    \end{tabularx}
\end{table}

\subsection{`Big' + `Dual' Configuration}
\label{`Big' + `Dual' Configuration}

\begin{figure}[H]
\begin{adjustwidth}{-\extralength}{0cm}
\centering
\includegraphics[width=17.5cm]{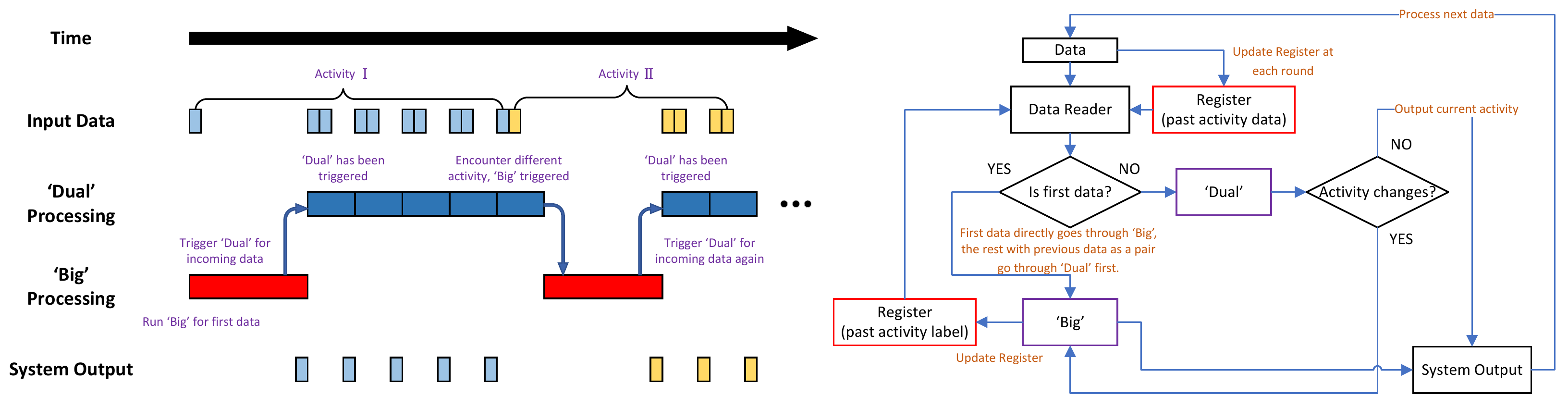}
\end{adjustwidth}
\caption{The processing steps (\textbf{left}) and the flow chart (\textbf{right}) for the `big' + `dual' configuration of the adaptive neural network system. In the left figure, the two input data blocks represent a pair of adjacent data samples required by the `dual' network. In the right figure, registers store the previous data and label for the current process in the `dual' network.\label{Config_1B1D}}
\end{figure}

The `big' + `dual' configuration is an alternative method of the adaptive neural network system. We replace the six `little' models with one small neural network called `dual'. Compared to the `big' + six `little' model, this one only consists of one primary and one secondary network model instead of one + six networks. In order to replace six `little' networks designed for six categories with only one `dual' network, the data sample for the previous activity is required to be stored in a register and compared with the current activity data sample as shown in Figure~\ref{Config_1B1D}. Then, the `dual' network can recognize these patterns to distinguish whether the current activity changes or not. For example, the first activity is classified as STANDING by the `big' network, and the second activity of SITTING is compared with the one previously stored by the `dual' network. If the `dual' network detects these two activities are not the same, the `big' network will be triggered for further inferences. Otherwise, the `dual' network keeps active for time and energy saving as shown in Figure~\ref{Config_1B1D}.

\begin{figure}[H]
\centering
\includegraphics[width=0.3\textwidth]{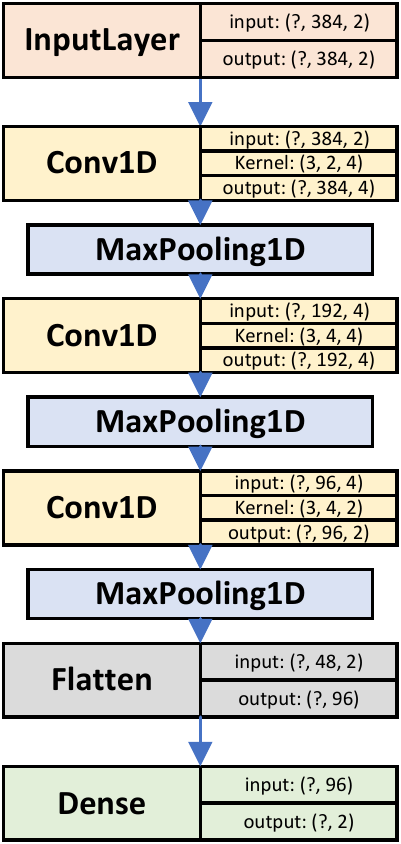}
\caption{`Dual' model structures in Keras.\label{NNStruc_Dual}}
\end{figure}

The `big' network is the same as the one introduced in the previous configuration, while the secondary `dual' network has been reconstructed as shown in Figure~\ref{NNStruc_Dual} and Table~\ref{Param_Dual}. In the same way as for the `little' network, the single input data from the total accelerometer sensor are selected for the `dual' network. Therefore, the input data shape of the `dual' network becomes (1, 128, 3, 2), which contains two adjacent input data samples. As there is a significant increase in the input data shape, the number of parameters increases from 184 in the `little' network to 300 in the `dual' network.

\vspace{-6pt}

    \begin{table}[H]
    \caption{`Dual' model parameter details. The pooling layers are hidden. For more info, see Figure~\ref{NNStruc_Dual}. \label{Param_Dual}}
        \begin{tabularx}{\textwidth}{CCC}
            \toprule
            \multicolumn{3}{c}{\textbf{Model: `Dual'}}\\
            \midrule
            \textbf{Layer (Type)} & \textbf{Output Shape} & \textbf{Param\#}\\
            \midrule
            model\_input & [(None, 384, 2)] & 0\\
            conv1d & (None, 384, 4) & 28\\
            conv1d\_1 & (None, 192, 4) & 52\\
            conv1d\_2 & (None, 96, 2) & 26\\
            model\_output & (None, 2) & 194\\
            \midrule
            \multicolumn{3}{c}{Total params: 300}\\
            \bottomrule
        \end{tabularx}
    \end{table}

\subsection{`Big' + Distance Configuration}
\label{`Big' + Distance Configuration}

\begin{figure}[H]
\begin{adjustwidth}{-\extralength}{0cm}
\centering
\includegraphics[width=17.5cm]{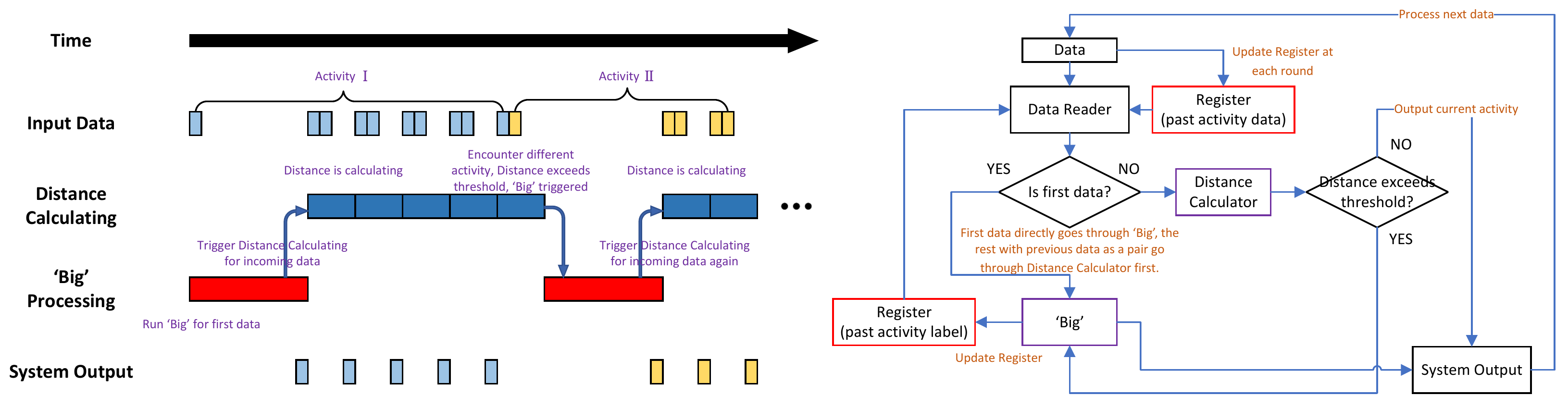}
\end{adjustwidth}
\caption{The processing steps (\textbf{left}) and the flow chart (\textbf{right}) for the `big' + distance configuration of the adaptive neural network system. In the left figure, the two input data blocks represent a pair of adjacent data samples required by the distance calculator. In the right figure, the registers store the previous data and label for the current process in the distance calculator.\label{Config_1BDD}}
\end{figure}

Finally, we consider whether the wake-up module in the adaptive system can be replaced by a simpler algorithm instead of using neural networks such as `little' and `dual' networks. This configuration, which is similar to the second configuration, replaces the `dual' network model with a distance calculator measuring the difference in the distance between two adjacent input samples. In order to pick up on an activity change, a distance calculator using Minkowski distance and Mahalanobis distance is applied to trigger the `big' network when the difference in distance reaches a pre-set threshold value as shown in Figure~\ref{Config_1BDD}. 
\begin{linenomath}
\begin{equation}
\label{equa_Minkowski}
D(x,y) = \left(\sum_{i=1}^{n} |x_i-y_i|^p\right)^{1/p}
\end{equation}
\end{linenomath}
\vspace{-6pt}

The Euclidean distance is a typical metric that measures the real distance of two points in N-dimensions. As shown in Equation~\eqref{equa_Minkowski}, Minkowski distance is a generalized format of Euclidean distance. When $ p=2 $, it becomes equivalent to the Euclidean distance, while it becomes equivalent to the Manhattan distance when $ p=1 $. Moreover, the Mahalanobis distance measures the distance of a target point P and a mean point of a distribution D. This distance increases if point P moves away along each principal component axis of D. The Mahalanobis distance becomes Euclidean distance when these axes are scaled to have a unit variance~\cite{de2000mahalanobis, mclachlan1999mahalanobis}.

The input data shape which is (1, 128, 3, 2) for the `dual' network should be stretched into (1,~384,~2) where the value two means that two adjacent data samples are required by the distance calculator. The calculator then measures the Minkowski distance between these two adjacent data samples following Equation~\eqref{equa_Minkowski} for both cases of $ p=1 $ and $ p=2 $. Mahalanobis distance requires the covariance matrix of the data set before the calculation. To wake up the `big' model, multiple thresholds can be selected to achieve multiple sensitivities. The `big' model is only triggered when the distance between the previous data sample and the current one is beyond the pre-set threshold. Therefore, a lower threshold value will reach a higher inference accuracy because the `big' network will be invoked more frequently. Conversely, a higher threshold value means that the `big' network is invoked fewer times, leading to a shorter inference time.

\section{Neural Network Microcontroller Deployment}
\label{Neural network Microcontroller Deployment}

The neural network models in the Keras format are quantized to the UINT8 format to reduce the amount of memory needed before MCU deployment. According to Equation~\eqref{equa_quantization} in~\cite{web_TFConverter}, as shown below, the real value is the input value of the training process in the range of [\textminus128, 127], while the quantized value is the target value after the quantization, which is in the UINT8 range of [0, 255]. The mean and the standard deviation values can be calculated as 128 and 1, respectively. Finally, the model in a quantized format is obtained. 
\begin{linenomath}
\begin{equation}
\label{equa_quantization}
real\_value = (quantized\_value - mean\_value)/std\_dev\_value
\end{equation}
\end{linenomath}

\begin{figure}[H]
\begin{adjustwidth}{-\extralength}{0cm}
\centering
\includegraphics[width=17.5cm]{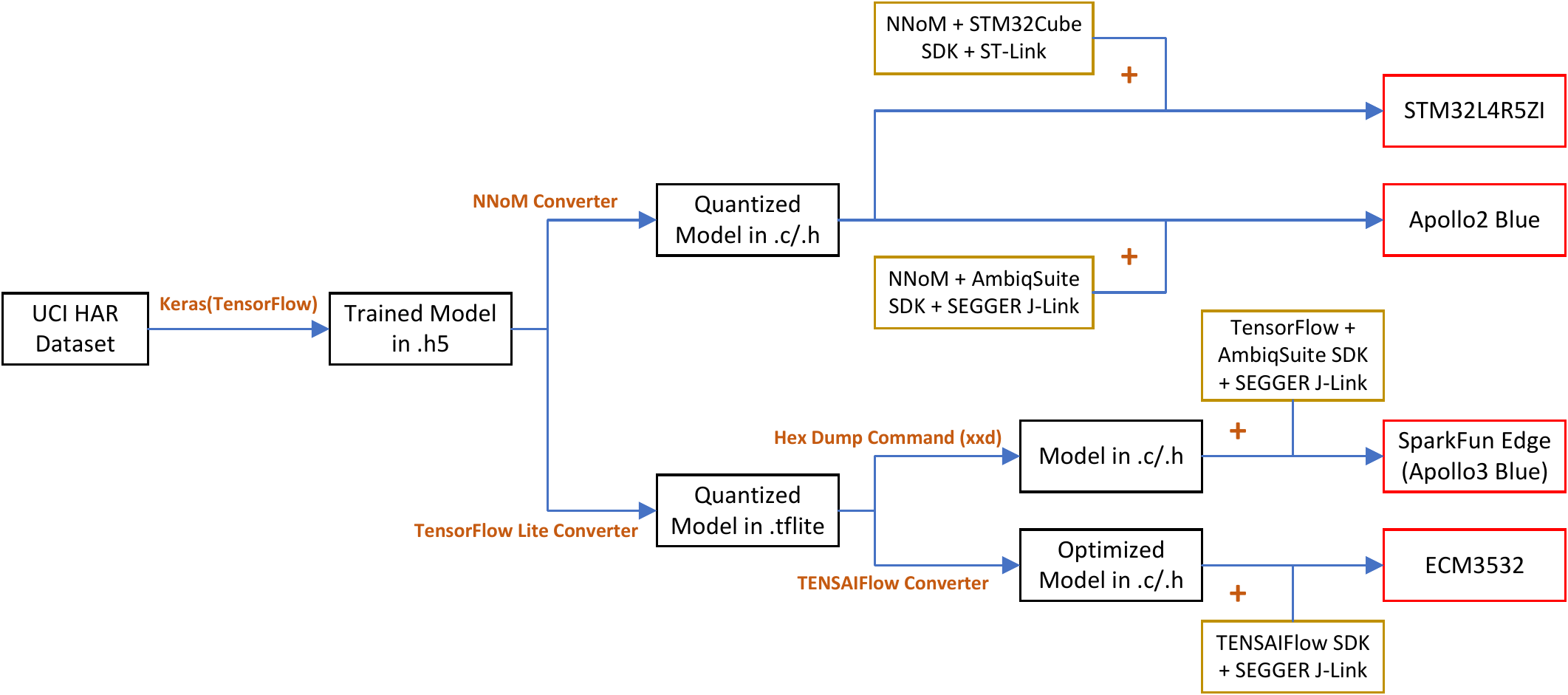}
\end{adjustwidth}
\caption{The map of the steps of neural network model deployment on target MCU boards. Black frames represent trained models, brown frames represent the library source codes used, while red ones represent MCU boards.\label{Deployment_chart}}
\end{figure}

We use the available data samples from UCI-HAR~\cite{web_UCI} instead of real-time data to perform a fair comparison across the different platforms. Thus, when the MCU runs the application, stored data and network models can be accessed correctly. Moreover, the model-switching algorithm for the adaptive system introduced in Section~\ref{Adaptive Neural Network Methodology} is achieved at the C code level instead of the network model layer level. The `big' and `little' models are capable of being invoked independently, which means the adaptive system is more flexible and effective at finding the balance between performance and energy consumption. Finally, before flashing the target boards, the application must be compiled to an executable binary using cross-compilation tools for GCC~\cite{web_GCC}, ARM Compiler~\cite{web_GNU} and the \textit{TENSAIFlow} compiler from Eta Compute~\cite{web_TENSAI}. The model deployment process is shown in Figure~\ref{Deployment_chart} and \ref{Deployment_table}.

\subsection{STM32L4R5ZI}
\label{STM32L4R5ZI}

\begin{figure}[H]
\begin{adjustwidth}{-\extralength}{0cm}
\centering
\includegraphics[width=17.5cm]{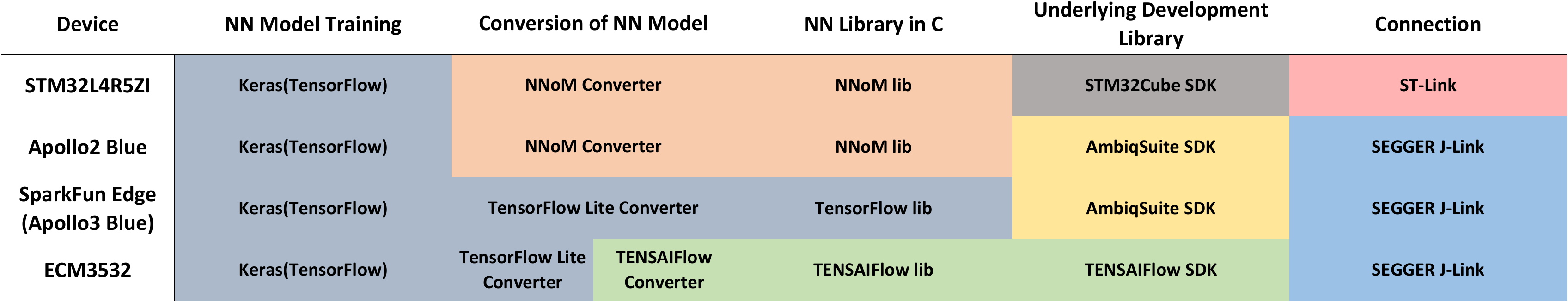}
\end{adjustwidth}
\sethlcolor{yellow}
\caption{Comparison
 of the software used in each deployment phase for different MCUs.\label{Deployment_table}}
\end{figure}

\textit{STM32Cube.AI} from STMicroelectronics~\cite{web_STM32} is a framework designed to optimize STM devices such as STM32L4. However, due to the limitation of being a proprietary environment, the switching algorithm between primary and secondary networks cannot be deployed at the C code level. On the other hand~\cite{web_NNoM}, it has been designed with a focus on general-purpose and flexible deployment on different MCU boards. The \textit{NNoM} converter is able to convert the pre-trained model in the Keras format to the C code and its neural network library can be used to deploy the model. Therefore, the \textit{NNoM} framework is selected for model deployment on STM32L4 instead of \textit{STM32Cube.AI} (see Figure~\ref{Deployment_table}).

\unskip

The STM32Cube SDK version 1.17.0 from STMicroelectronics which contains utility tools and example projects, is required to drive the STM32L4R5ZI MCU board. Keil uVision IDE from ARM is chosen to set up a coding environment to support STM32L4. The driver pack for STM32L4 is required to be installed by the pack installer of Keil. The STM32L4 CN1 port is connected with a desktop PC by using a micro-USB cable. Then, the ST-Link debugger can be selected under the target debug page and the STM32L4 device can then be connected and detected by the PC. Alternatively, if the connection is unsuccessful, STM32 ST-LINK Utility from STMicroelectronics can erase the board to avoid software conflicts. 

After the NN models are trained by Keras (\textit{TensorFlow}\_v1.15), they are required to be quantized by applying the \textit{NNoM} converter command as shown in Listing~\ref{Command_NNoM}. Then, the header file containing model weights can be generated by using the function below. Before building the project, the weight header file, input data file and the files from the \textit{NNoM} library should be added by Keil Manage Project Items. Finally, the steps of building and flashing the project to the development board can be carried out. To observe the output from the debug viewer, the core clock under the trace page of the target debug setting should match the operating clock of the device.

\begin{lstlisting}[caption={NNoM converter Python function for quantized model generation.}, label=Command_NNoM]
generate_model(model, x_test, name=`weight.h')
\end{lstlisting}

\subsection{Apollo2 Blue}
\label{Apollo2 Blue}

AmbiqSuite SDK version 2.2.0 from Ambiq supports the model deployment on Apollo2 Blue. Keil uVision IDE from ARM is used to set up a coding environment. After installing the driver pack for Apollo2, the Apollo2 board is connected to the PC by using a micro-USB cable and selecting J-Link under the target debug page of Keil as shown in Figure~\ref{Deployment_chart}. Similar to the case of STM32L4, Apollo2 is not supported by \textit{TensorFlow Lite} and \textit{TENSAIFlow}. Thus, the pre-trained models in Keras format are converted into a quantized format using the \textit{NNoM} converter as shown in Listing~\ref{Command_NNoM}. Then, the model weights and data header files and the \textit{NNoM} library should be added into the project by Keil Manage Project Items. After building and flashing the project to the target board, the Keil debug viewer can be used to observe the model outputs.

\subsection{SparkFun Edge (Apollo3 Blue)}
\label{SparkFun Edge (Apollo3 Blue)}

\textit{TensorFlow} from Google is not only capable of training neural network models, but also includes \textit{TensorFlow Lite} to deploy network models on edge devices such as MCUs~\cite{web_TFLite}. The trained network model saved in the Keras format can be converted into the quantized format using the \textit{TensorFlow Lite} converter in Listing~\ref{Command_TFLite_B} and \ref{Command_TFLite_L}. The library source code in C and board SDK files are provided to support the model deployment on MCUs (see Figure~\ref{Deployment_chart}). We use \textit{TensorFlow Lite} to support model deployment for the MCU development board of SparkFun Edge (Apollo3).

AmbiqSuite SDK version 2.2.0 contains utility tools and drivers from Ambiq to support SparkFun Edge (Apollo3 Blue). \textit{TensorFlow Lite} version 1.15 is used to convert Keras models using floating-point parameters into the TFLite model with UINT8 parameters. As per the corresponding command lines in Listing~\ref{Command_TFLite_B} and \ref{Command_TFLite_L}, the quantized model files are generated and ready to be deployed. The TFLite model is converted into a hexadecimal file which can be read by the \textit{TensorFlow Lite} library by using hex dump command `xxd'. Finally, we connect Apollo3 to the PC with a micro-USB cable and flash the binary file to the target board using the flash utility provided by the AmbiqSuite SDK.

\begin{lstlisting}[caption={TensorFlow Lite converter command lines for `big' model quantization.}, label=Command_TFLite_B]
tflite_convert \
    --keras_model_file=./Output_Models/${MODELNAME}.h5 \
    --output_file=./Output_Models/${MODELNAME}.tflite \
    --inference_type=QUANTIZED_UINT8 \
    --input_shapes=1,128,3:1,128,3:1,128,3 \
    --input_arrays=model_input1,model_input2,model_input3 \
    --output_arrays=model_output/BiasAdd \
    --default_ranges_min=0 --default_ranges_max=255 \
    --mean_values=128,128,128 --std_dev_values=1,1,1 \
    --change_concat_input_ranges=false \
    --allow_nudging_weights_to_use_fast_gemm_kernel=true \
    --allow_custom_ops
\end{lstlisting}

\begin{lstlisting}[caption={TensorFlow Lite converter command lines for `little' model quantization.}, label=Command_TFLite_L]
tflite_convert \
    --keras_model_file=./Output_Models/${MODELNAME}.h5 \
    --output_file=./Output_Models/${MODELNAME}.tflite \
    --inference_type=QUANTIZED_UINT8 \
    --input_shapes=1,128,3 \
    --input_arrays=model_input \
    --output_arrays=model_output/BiasAdd \
    --default_ranges_min=0 --default_ranges_max=255 \
    --mean_values=128 --std_dev_values=1 \
    --change_concat_input_ranges=false \
    --allow_nudging_weights_to_use_fast_gemm_kernel=true \
    --allow_custom_ops
\end{lstlisting}

\subsection{ECM3532}
\label{ECM3532}

\textit{TENSAIFlow} from Eta Compute is a framework designed to deploy pre-trained network models for Eta products such as ECM3531 and ECM3532~\cite{web_TENSAI}. It is highly optimized for Eta Compute products to achieve the best balance between performance and efficiency. This framework is not capable of training neural network models such as \textit{TensorFlow}; it only provides the model conversion and deployment after training. After the pre-trained model is converted into a quantized TFLite format by \textit{TensorFlow Lite}, \textit{TENSAIFlow} converts the TFLite model to the C code which can be invoked with the library source code. The ECM3532 development board is not supported by \textit{NNoM} or \textit{TensorFlow Lite}. Therefore, \textit{TENSAIFlow} SDK version 2.0.2 contains the \textit{TENSAIFlow} converter and neural network library from Eta Compute required to support model deployment on ECM3532. As shown in Figure~\ref{Deployment_chart}, the pre-trained model from Keras (\textit{TensorFlow}\_v1.15) is quantized to the UINT8 format by using the \textit{TensorFlow Lite} converter first (Listing~\ref{Command_TFLite_B} and \ref{Command_TFLite_L}) and converted to the readable format for the \textit{TENSAIFlow} library by using the \textit{TENSAIFlow} converter (Listing~\ref{Command_TENSAIFLow}). Then, we build the project and flash it to the target ECM3532.

\begin{lstlisting}[caption={TENSAIFlow converter command lines for model conversion.}, label=Command_TENSAIFLow]
./tensaiflow_compile \
    --tflite_file ../model_zoo/${MODELNAME}.tflite \
    --out_path ../../../Applications/${PROJECTNAME}/src/ \
    --weights_path ../../../Applications/${PROJECTNAME}/include/
\end{lstlisting}

\section{Results and Discussion}
\label{Results and Discussion}

\begin{figure}[H]
\includegraphics[width=\textwidth]{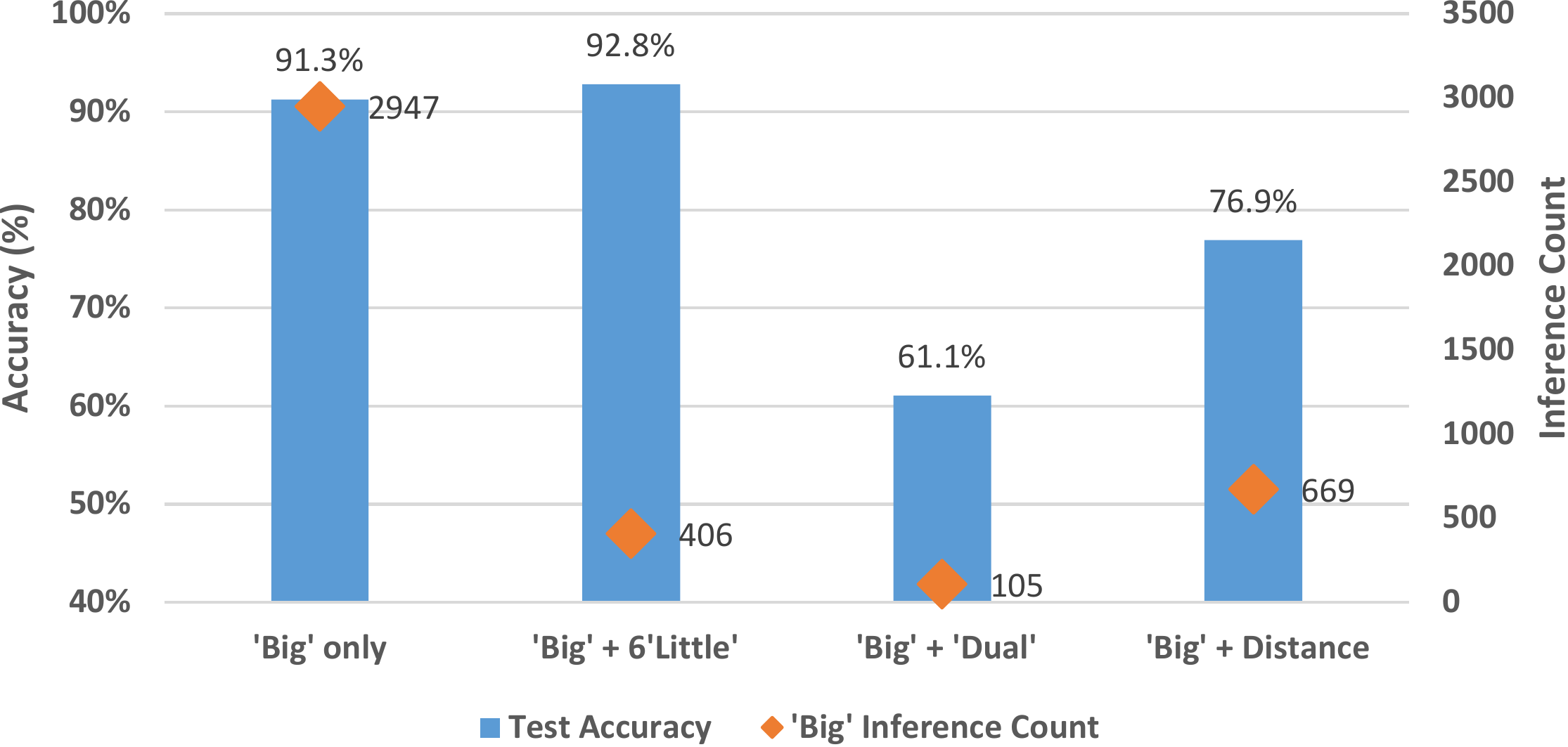}
\caption{The accuracy and the `big' inference counts for four configurations (quantized TFLite format) on the test data set (2947 samples) have been evaluated on a PC.\label{PC}}
\end{figure}

The accuracy of the different configurations and the original using the full HAR test data set is shown in Figure~\ref{PC}. We do not consider different random initialization seeds in this work but we use the same trained network for the different MCUs to perform a fair comparison. We also choose the learning rate carefully, using a relatively slow rate and SGDR to prevent the model from sinking into a locally optimal point instead of the global one. We use holdout cross validation to divide the whole data set: 70\% for the training data set, 15\% for the validation data set and 15\% the testing data set.

In Figure~\ref{PC}, the `big'-only configuration has 91.3\% accuracy but the model has a large invocation count that will result in significant latency. The `big' + six `little' configuration reaches a comparable level of accuracy and the number of times the `big' model invoked is reduced from 2947 to 406, reducing the inference time of the `big' model by two-thirds. The `big' + `dual' configuration cannot reach a similar accuracy due to the low accuracy of the secondary `dual' network. The `big' + distance configuration achieves a relatively low testing accuracy and it invokes the `big' network 669 times in 2947 data samples.

In order to establish the same testing environments and provide the same test samples for all MCU boards, we choose to apply the data samples from the UCI-HAR test data set rather than a real-time signal from the board sensors. Only 60 data samples from the UCI-HAR test data set can be selected to fit in the MCU boards together with the model and switching algorithm due to memory limitations. Therefore, we select ten data samples for each activity and compose them into a certain sequence of activity \uppercase\expandafter{\romannumeral1} to \uppercase\expandafter{\romannumeral6}. This means that there are five activity changes in the test data sequence. We have verified that the classification results obtained in these 60 samples are equivalent to the ones obtained with the whole data set, although there are some negligible differences between devices due to the different toolchains. The following evaluations are performed under a working frequency of 48~MHz without debugging. Four configurations of the adaptive neural network are evaluated below:

\subsection{`Big' Only}
\label{`Big' only}

\begin{figure}[H]
\centering
\includegraphics[width=\textwidth]{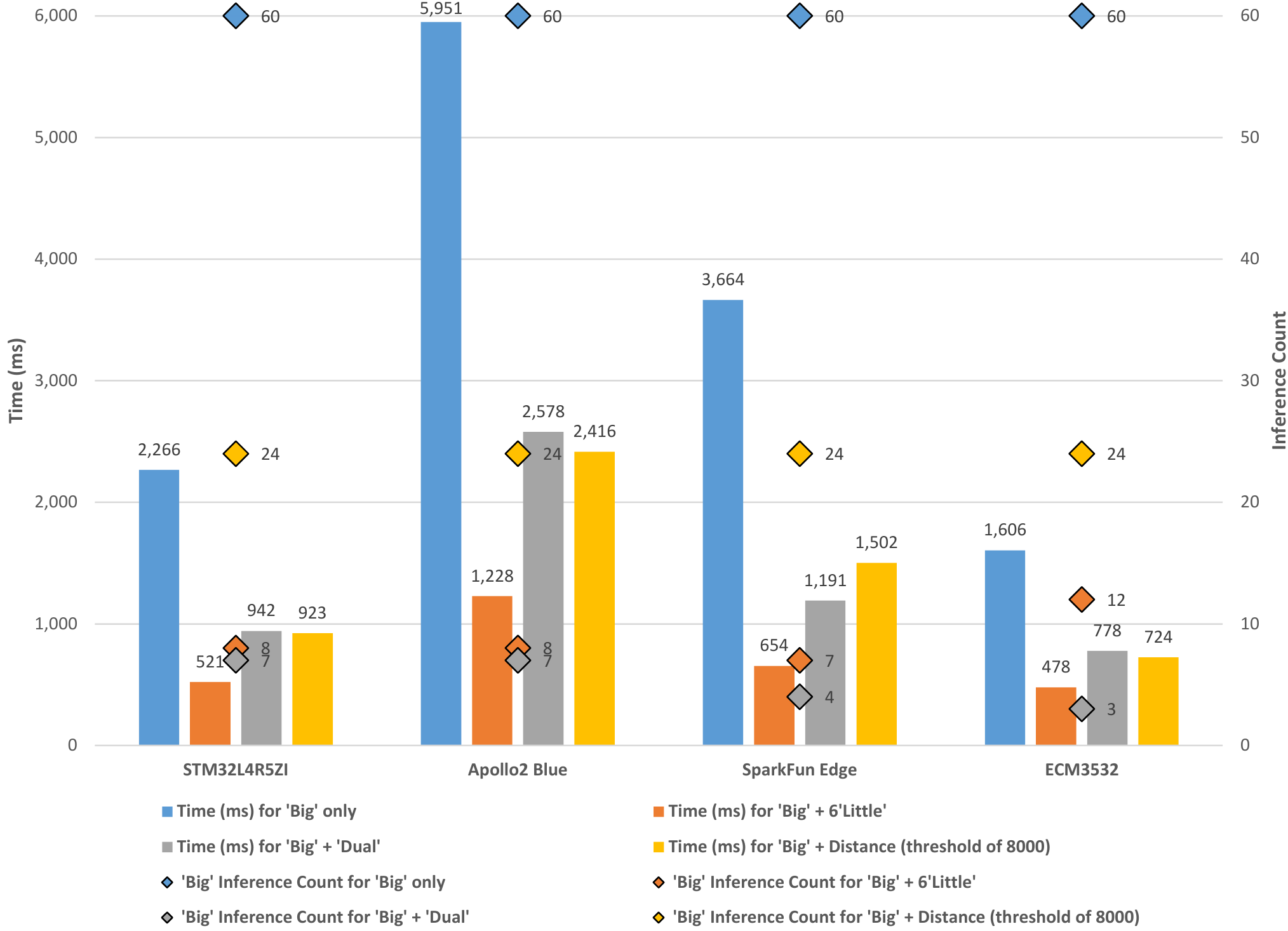}
\sethlcolor{red}
\caption{The time evaluation of four adaptive configurations on MCU boards with the `big' inference counts. A total of 60 data samples extracted from the UCI-HAR test data set are tested to form the evaluation.\label{MCU_Time}}
\end{figure}

\begin{figure}[H]
\centering
\includegraphics[width=\textwidth]{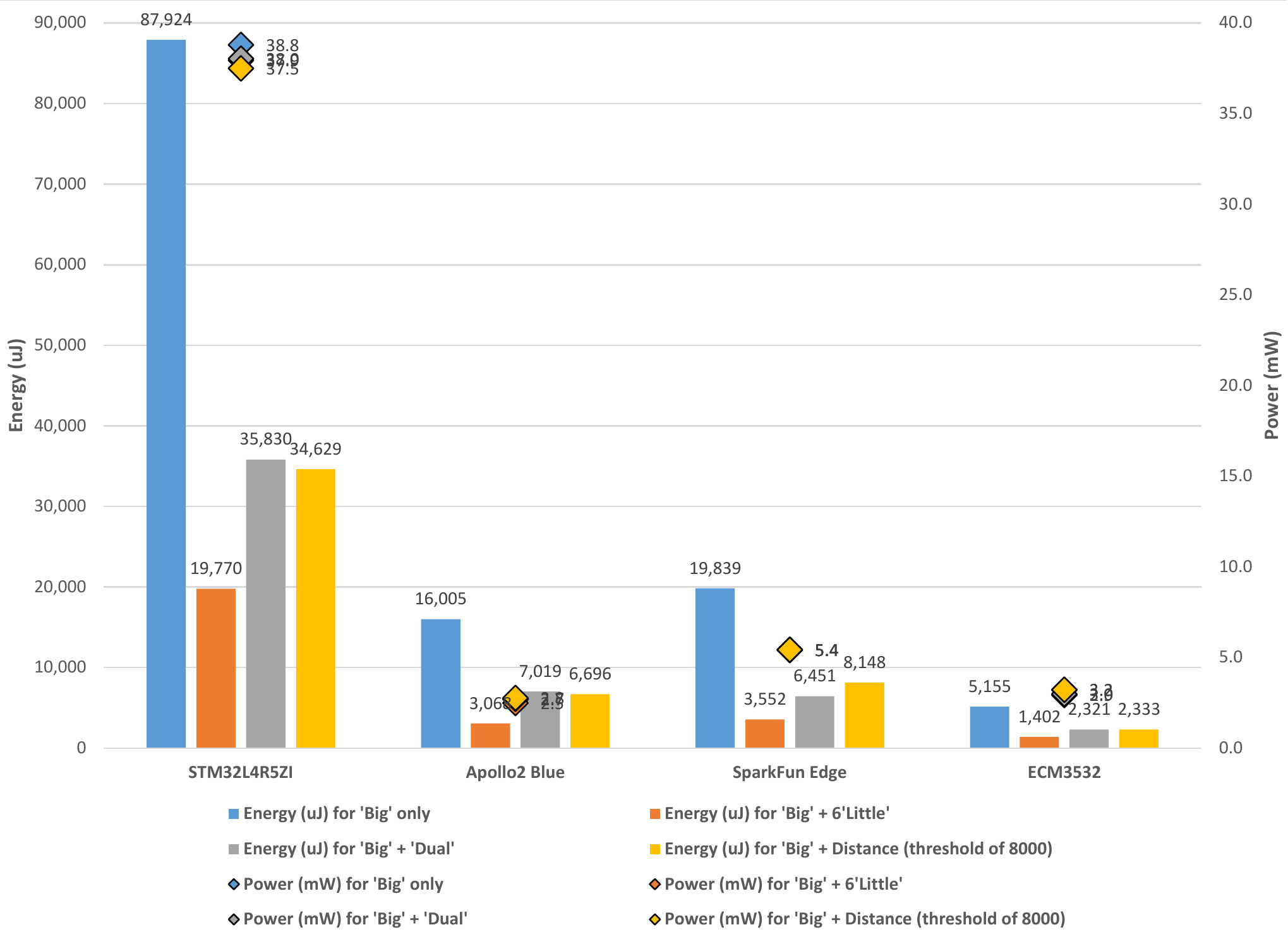}
\sethlcolor{red}
\caption{The power and energy evaluation of four adaptive configurations on MCU boards. A total of 60 data samples extracted from the UCI-HAR test data set are tested to form the evaluation.\label{MCU_Energy}}
\end{figure}

As shown in Figure~\ref{PC}, after removing the LSTM layers used in~\cite{ordonez2016deep}, we still maintain an accuracy level of around 90\% for the `big' network on the activity classification task compared to the results in~\cite{nunez2021energy,ordonez2016deep}. The original `big' model method performs 2947 inferences on all test data samples. Due to the large topology of the `big' model and a large number of inferences, the `big'-only configuration has the highest execution time. This can be seen in Figure~\ref{MCU_Time}: for all four MCUs working at the same operating frequency of 48~MHz, the latency of the `big'-only model is the highest among all four configurations. The power consumption values for each configuration show negligible variations for each MCU in Figure~\ref{MCU_Energy}. Therefore, the energy consumption for each configuration is only affected by the inference time. As shown in Figure~\ref{MCU_Energy}, the `big'-only configuration consumes the highest value of energy.

\unskip

\subsection{`Big' + Six `Little'}
\label{`Big' + Six `Little'}

The difference between the inference time of the `big'-only and `big' + six `little' configurations is shown in Figure~\ref{MCU_Time}. The inference latency of the `big' model is around 12 times longer than the latency of the `little' model. Therefore, the lower the number of times the `big' network gets invoked, the higher the efficiency of the system is. In this configuration, six `little' models are applied to save time by restricting the `big' inference count to around ten times. In all MCU evaluations in Figure~\ref{MCU_Time}, the time result of the `big' + six `little' configuration is the lowest and this reduces the execution time by around 80\% compared to the original `big'-only configuration and around 50\% compared to the others. For all four configurations, the power is largely equivalent, as can be seen in Figure~\ref{MCU_Energy}. Due to the significant advantage of the `big' + six `little' configuration in terms of execution time, this configuration achieves energy savings of around 80\% compared to the original `big' method on all MCUs.

\subsection{`Big' + `Dual'}
\label{`Big' + `Dual'}

\begin{figure}[H]
\includegraphics[width=\textwidth]{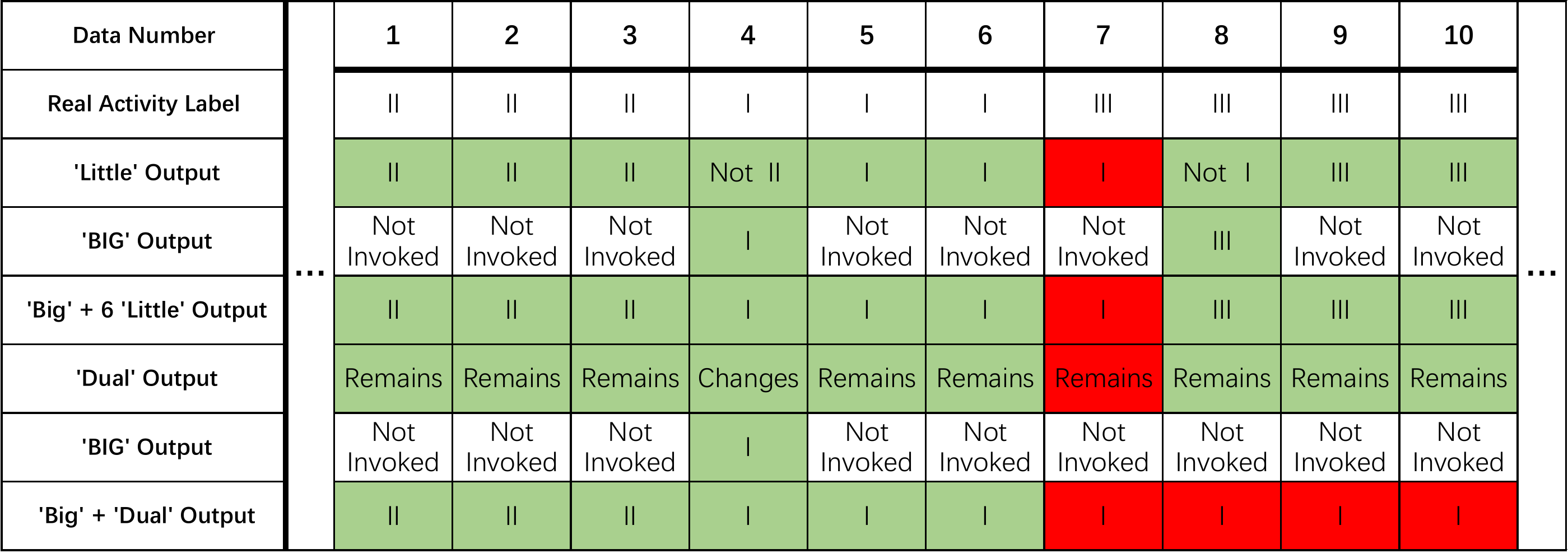}
\caption{The output comparison of two configurations when an error occurs at the moment of an `activity change'. Errors have been labelled in the color red. The results including the primary module, secondary module, and overall adaptive system have been shown below.\label{Error_1B1D}}
\end{figure}

In contrast to the `big' + six `little' configuration, the `big' + `dual' configuration is not restricted by the number of categories that need to be classified. The number of `little' networks in the previous configuration is determined by the number of categories, which leads to difficulties in model deployment if the number of categories is large such as in the CIFAR-100 data set. By applying a network focusing on detecting activity changes, the `big' + `dual' configuration can pick up activity changes by comparing the current activity and the previous activity. However, two deficiencies appear in this configuration. Firstly, in the 7352 training data samples, there are only 280 cases of activity switching. We extract 280~data samples with an `activity change' label and 7072 samples with an `activity continuance' label to train the `dual' model, resulting in an unbalanced training data set. Secondly, there is an error propagation problem which occurs when `dual' classification is incorrect in the case of `activity change'. For example, in Figure~\ref{Error_1B1D}, the `dual' model has an error at the seventh data sample where the activity switches from \uppercase\expandafter{\romannumeral1} to \uppercase\expandafter{\romannumeral3}, skipping `big' inference and misleading the adaptive system to output activity \uppercase\expandafter{\romannumeral1}. After that, the `dual' model has no errors for the rest of the data, detecting no activity changes. This adaptive system continues to propagate the output errors because the seventh output is set up as activity \uppercase\expandafter{\romannumeral1} instead of \uppercase\expandafter{\romannumeral3}. Compared to the `big' + six `little' configuration, the `big' model is also skipped at the seventh data because the `little' model does not pick up any changes (an error). However, after the next data input, the `little' model is able to recognize that the activity is not activity \uppercase\expandafter{\romannumeral1} anymore. Then, the `big' model is invoked to output the correct activity label and the system recovers to a correct state. 

\textls[-10]{Although the `dual' model is able to solve the large category issue, it is not sufficiently trained due to the unbalanced training data set. Due to the poor accuracy of the `dual' model, the error propagation mentioned in Figure~\ref{Error_1B1D} occurs and fails to switch on the `big' model when detecting an activity change for further inference. This results in a minimal `big' inference count but a relatively poor performance in terms of accuracy. Therefore, the overall accuracy of this adaptive system (around 60\% for all test data on a PC) is lower compared to the other configurations as shown in Figure~\ref{PC}. Furthermore, because the complexity of the `dual' model is relatively high and the `dual' model is activated continuously, this leads to a higher complexity in the inference process. Additionally, the combination of previous and current data samples for the `dual' input needs to be pre-processed. Therefore, despite having the fewest number of `big' inference counts (Figure~\ref{PC}), the latency and energy consumption double compared to the best configuration of the `big' + six `little' models as shown in Figures~\ref{MCU_Time} and~\ref{MCU_Energy}.}

\subsection{`Big' + Distance}
\label{`Big' + Distance}

\begin{figure}[H]
\includegraphics[width=\textwidth]{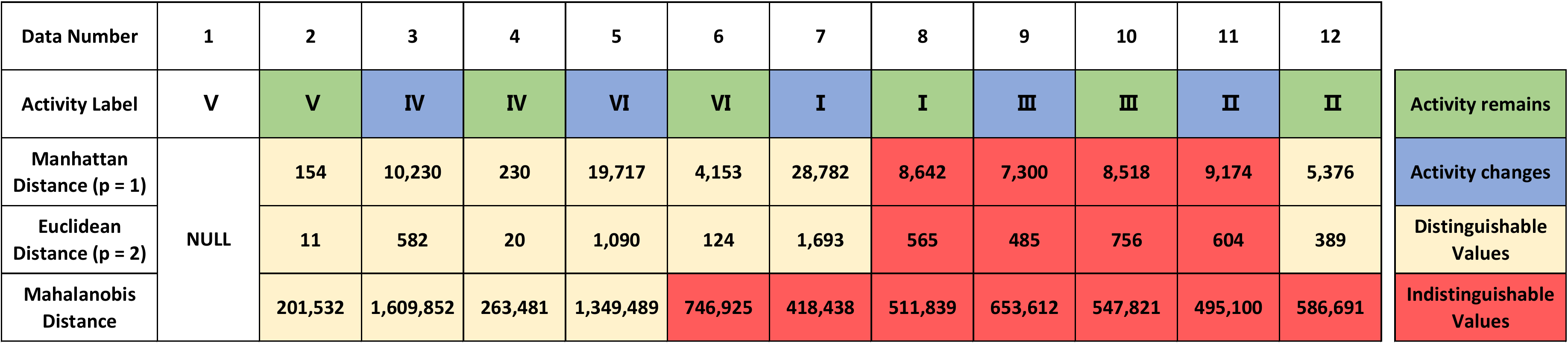}
\sethlcolor{red}
\caption{Distance results for different pairs of adjacent data samples. Values in one distance measurement (one row) are comparable.\label{Distance_1BDD}}
\end{figure}

The `big' + distance configuration, as shown in Figure~\ref{Distance_1BDD}, shows that the Manhattan distance and Euclidean distance have a poor performance when distinguishing activities \uppercase\expandafter{\romannumeral1} to \uppercase\expandafter{\romannumeral3} which are WALKING, WALKING\_UPSTAIRS, and WALKING\_DOWNSTAIRS. The distance between the data samples of the same activities exceeds the distance between the ones of different activities (see data 8 to 10 in Figure~\ref{Distance_1BDD}). Therefore, a clear threshold boundary cannot be set to separate the case of `activity change' from unchanged activities due to these indistinguishable values.

In the `big' + distance configuration, a threshold point of 8000 for the Manhattan distance is selected for the evaluation in Figures~\ref{PC} and~\ref{MCU_Time}. This threshold of 8000 triggers the `big' model more frequently so it can be considered sensitive. As with the `big' + `dual' model, the `big' + distance model also suffers from the error propagation issue which severely affects the overall accuracy. Compared to the `big' + six `little' configuration, this configuration achieves a relatively low accuracy level at around 76\% with a higher number of `big' invocation times as shown in Figure~\ref{PC}. Furthermore, this configuration has a significant latency and energy costs which doubles compared to `big' + six `little' models and it is similar to the `big' + `dual' configuration as shown in Figures~\ref{MCU_Time} and~\ref{MCU_Energy}.

Overall and across all MCUs, our best adaptive network configuration, the `big' + six `little' configuration, achieves a high prediction accuracy level of around 90\%, which is comparable to the original `big'-only method. As discussed in Section~\ref{Low-power Microcontroller Evaluation}'s initial evaluation of the MCU, ECM3532 achieves the highest processing speed, followed by STM32L4, SparkFun Edge (Apollo3) and Apollo2 (listed fastest to slowest). With the same configuration, the execution time in Figure~\ref{MCU_Time} shows that this is consistent across all four MCU boards. For the `big' + six `little' configuration, the `big' inference count is reduced by around 85\% compared with the original method, achieving up to $5\times$ the acceleration on MCUs. Since the MCU boards are in working mode when running different configurations, the power consumption of these configurations is similar to the MCU shown in Figure~\ref{MCU_Energy}. Due to the negligible differences between network configurations in terms of power, the distribution of the energy consumption of the configurations for each MCU follows the time cost distribution in Figure~\ref{MCU_Time}. As shown in Figure~\ref{MCU_Energy}, across all devices, the `big' + six `little' algorithm configuration achieves energy savings of around 80\% compared to the original `big'-only method, and around 50\% compared to the other two configurations. Furthermore, compared to a standard MCU running the ‘big’ network only, the best configuration, the `big' + six `little' model, coupled with the best state-of-the-art near-threshold hardware, can achieve a reduction in energy of up to 98\% that will translate into a $62\times$ increase in the operating lifetime of an application for detecting battery-powered activity.

\section{Conclusions}
\label{Conclusions}

In this research, we have compared commercially available near-threshold and standard MCUs in terms of performance and energy consumption. At the same operating frequency, near-threshold MCUs have a significant advantage in power consumption, which is 80\% lower than standard MCUs. Due to the comparable processing speed, the low-power near-threshold MCU can achieve energy savings of around 80\% compared to the standard MCU STM32L4. Moreover, we have proposed three adaptive neural network configurations and investigated how MCU deployments can benefit from our algorithms to obtain lower energy consumption while maintaining prediction accuracy. We demonstrate that despite the low amount of memory available in these devices, it is possible to deploy `big--little' configurations that result in significantly better energy and performance characteristics. The proposed algorithms can be successfully deployed on STM32L4R5ZI, Apollo2 Blue, SparkFun Edge (Apollo3 Blue) and ECM3532. The application UCI-HAR is representative of an activity recognition task that assumes that an activity will remain constant for some period of time before switching to a different activity. In order to save time and energy, we activate the secondary model with a faster inference speed to pause the primary model when the activity remains constant. The best adaptive network configuration, the `big' + six `little' configuration, has achieved a reduction in energy of 80\% and a comparable level of prediction accuracy to the original method in the UCI-HAR test. The results prove that the proposed methods can deliver different levels of time--energy reduction and constant accuracy on all the devices we tested. Furthermore, coupled with near-threshold MCUs, the best configuration is able to increase battery life by up to 62x on UCI-HAR compared to the original non-adaptive method using a standard MCU.

Future work involves extending the work to other application areas such as machine health monitoring and anomaly detection. In addition, we plan to investigate how the approach can be scaled to applications with a large number of possible output categories without an explosion in the memory requirements by using additional network hierarchies. Finally, a future research direction includes developing a framework that is able to automatically extract optimal `little' configurations from a `big' configuration in terms of overall accuracy and energy in order to replace manual analysis.

\vspace{6pt}

\authorcontributions{Methodology, Z.S., N.H. and J.N.-Y.; software, Z.S. and J.N.-Y.; validation, Z.S.; resources, Z.S.; data curation, Z.S.; writing---original draft preparation, Z.S.; writing---review and editing, Z.S., N.H. and J.N.-Y.; visualization, Z.S.; supervision, J.N.-Y.; All authors have read and agreed to the published version of the manuscript.}

\funding{This work was partially funded by the Royal Society INF/R2/192044 Machine Intelligence at the Network Edge (MINET) fellowship.}

\institutionalreview{Not applicable}

\informedconsent{Not applicable}

\dataavailability{Our work can be found here: (accessed on 9 March 2022) \url{https://github.com/DarkSZChao/Big-Little_NN_Strategies}.} 


\conflictsofinterest{The authors declare no conflicts of interest.}

\abbreviations{Abbreviations}{
The following abbreviations are used in this manuscript:\\

\noindent 
\begin{tabular}{@{}ll}
MCU & Microcontroller Unit\\
LoT & Internet of Things\\
CNN & Convolutional Neural Network\\
UCI-HAR & UCI-Human Activity Recognition
\end{tabular}}

\begin{adjustwidth}{-\extralength}{0cm}

\reftitle{References}

%


\end{adjustwidth}
\end{document}